\documentclass[11pt]{article}

\usepackage[final]{acl}

\usepackage{times}
\usepackage{latexsym}

\usepackage{rotating}   
\usepackage{ragged2e}   
\usepackage[T1]{fontenc}

\usepackage[utf8]{inputenc}

\usepackage{microtype}

\usepackage{inconsolata}
\usepackage{bbm}
\usepackage{siunitx}  
\usepackage{times}
\usepackage{latexsym}
\usepackage{xcolor} 
\usepackage{amsmath} 
\usepackage{graphicx}
\usepackage{booktabs, multirow} 

\usepackage[linesnumbered,ruled]{algorithm2e} 

\usepackage{CJKutf8}
\usepackage{CJK}

\definecolor{lightred}{rgb}{1, 0.878, 0.741} 
\definecolor{lightteal}{rgb}{0.702, 0.851, 0.851}
\definecolor{lightyellow}{rgb}{1, 1, 0.8}

\usepackage{xcolor}
\usepackage{colortbl}
\usepackage{longtable}
\usepackage[skins,xparse,breakable]{tcolorbox} 
\usepackage{microtype}
\usepackage{tabularray}

\usepackage{tabularx}

\usepackage{placeins}

\newlength{\myMheight}
\newcommand{\valueWithStd}[2]{#1$_{\pm#2}$}
\newcommand{\valueWithoutStd}[2]{#1}

\title{MeasHalu: Mitigation of Scientific Measurement Hallucinations for Large Language Models with Enhanced Reasoning}

\author{
 \textbf{Ruijun Huang\textsuperscript{1}},
 \textbf{Zhiqiao Kang\textsuperscript{1}},
 \textbf{Yuxuan Zhu\textsuperscript{1}},
 \textbf{Junxiong Li\textsuperscript{1}},
 \textbf{Jiahao Zhao\textsuperscript{1}},
\\
 \textbf{Minghuan Tan\textsuperscript{1}\thanks{Corresponding author}},
 \textbf{Feng Jiang\textsuperscript{2}\footnotemark[1]},
 \textbf{Min Yang\textsuperscript{1}}
\\
\\
 \textsuperscript{1}Shenzhen Key Laboratory for High Performance Data Mining, \\Shenzhen Institutes of Advanced Technology, Chinese Academy of Sciences,\\
 \textsuperscript{2} Artificial Intelligence Research Institute, Shenzhen University of Advanced Technology
\\
 \small{
    \textbf{Correspondence:} \href{mailto:mh.tan@siat.ac.cn,jiangfeng@suat-sz.edu.cn}{mh.tan@siat.ac.cn, jiangfeng@suat-sz.edu.cn}
 }
}

\definecolor{lightblue}{RGB}{0,60,255}   
\begin{document}
\maketitle
\begin{CJK*}{UTF8}{gbsn}
\begin{abstract}

The accurate extraction of scientific measurements from literature is a critical yet challenging task in AI4Science, enabling large-scale analysis and integration of quantitative research findings. 
However, Large Language Models (LLMs) frequently exhibit severe hallucinations,
which significantly undermine the reliability of automated scientific document understanding systems. 
To address this problem, we propose \textsc{\textbf{MeasHalu}}, a novel framework for mitigating scientific measurement hallucinations through enhanced reasoning and targeted optimization. 
We first present a fine-grained taxonomy of measurement-specific hallucinations, categorizing errors across quantities, units, modifiers, and relations. 
Our approach incorporates a two-stage reasoning-aware fine-tuning strategy using augmented scientific data and process-based supervision. 
Furthermore, we introduce a progressive reward curriculum designed to penalize specific hallucination types, significantly improving extraction faithfulness.
Experimental results demonstrate that \textsc{MeasHalu} substantially reduces hallucination rates and improves overall accuracy on the MeasEval benchmark. 
This work provides a targeted solution to a key bottleneck in automated scientific knowledge extraction, facilitating more trustworthy and scalable machine-assisted scientific literature analysis. Our codes and data are publicly available on \href{https://github.com/CAS-SIAT-XinHai/MeasHalu}{https://github.com/CAS-SIAT-XinHai/MeasHalu}.
\end{abstract}

\begin{figure}[t]  
    \centering
    \includegraphics[width=0.8\linewidth]{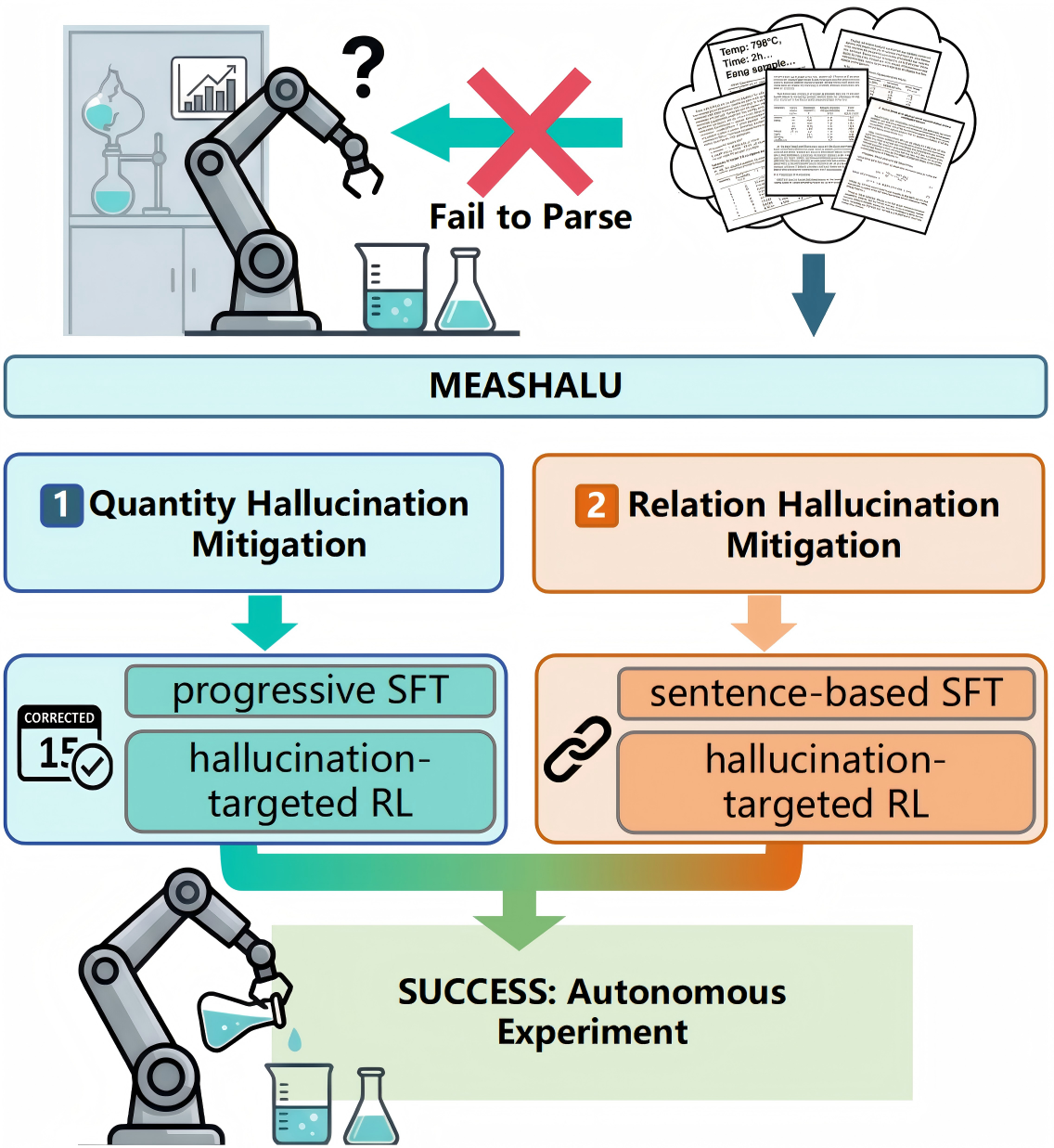} 
    \caption{\textbf{Motivation of MeasHalu.} To rectify parsing failures, we propose a taxonomy-based approach to mitigate quantity and relation hallucinations. }
    \label{fig:motivation}
\end{figure}

\section{Introduction}

The rapid expansion of scientific literature has created an unprecedented demand for reliable automatic extraction of quantitative knowledge, which lies at the core of modern AI4Science applications such as large-scale meta-analysis, knowledge base construction, and autonomous scientific discovery~\citep{10.1162/qss_a_00327,chen2025ai4researchsurveyartificialintelligence}. 
Central to this process is \emph{scientific measurement extraction}—the task of identifying numerical quantities, their units, modifiers, and their relationships to measured entities and properties. 
These quantitative statements form the evidential backbone of experimental sciences across disciplines ranging from materials science to biomedical research~\citep{berrahou2013extract,Kononova2021}. 
Although recent Large Language Models (LLMs) have demonstrated remarkable generalization abilities, they continue to perform unreliably on this task~\citep{foppiano2024mining}: even minor hallucinations in quantities or relations can invalidate entire experimental conclusions, severely limiting the trustworthiness of LLM-driven scientific understanding systems.

A key challenge underlying this failure is that \emph{measurement hallucinations differ fundamentally from general textual hallucinations}. 
Unlike open-domain factual errors, measurement hallucinations exhibit fine-grained structural failures: models fabricate nonexistent values, misassociate quantities with wrong entities, overlook crucial qualifiers, or distort relations between scientific variables~\citep{HyperPIE}. 
Existing hallucination mitigation techniques, such as retrieval augmentation~\citep{lewis2020retrieval}, generic instruction tuning, or conversational verification~\citep{polak2024extracting}, remain insufficient, as they are not designed to enforce the strict grounding and structural consistency required by scientific measurements. 
Yet, despite the importance of this problem, current research lacks both a systematic analysis of measurement-specific hallucination phenomena and targeted learning mechanisms for their mitigation. 
For instance, even state-of-the-art LLM-based extraction systems often compromise faithfulness by generating implicit information that is absent from the original text, such as inferring chemical formulas~\citep{dagdelen2024structured}.

In this work, we present \textsc{MeasHalu}, a reasoning-enhanced framework that explicitly models and suppresses scientific measurement hallucinations in LLMs. 
Our central insight is that hallucinations in this domain arise from two intertwined sources: 
(1) unreliable quantitative reasoning that corrupts individual quantities and units, and 
(2) fragile long-range relational reasoning that breaks the alignment between quantities, entities, and scientific properties. 
\textsc{MeasHalu} addresses these failure modes through a unified learning pipeline that combines reasoning-aware supervised fine-tuning with targeted reinforcement learning via structured reward shaping, thereby internalizing scientific grounding constraints directly into model parameters.

Concretely, \textsc{MeasHalu} introduces a fine-grained taxonomy of measurement hallucinations, and leverages this analysis to design a progressive optimization strategy: 
an initial supervised stage that standardizes quantitative reasoning and extraction structure, followed by Group Relative Policy Optimization (GRPO) with carefully constructed rewards that penalize fabrication, out-of-scope predictions, misclassification, and relational incompleteness. 
Our framework is developed on top of the MeasEval annotation schema~\citep{harper-etal-2021-semeval} and integrates external quantity validators, including CQE~\citep{almasian2023cqe} and Quantulum\footnote{\url{https://github.com/nielstron/quantulum3}}, during training. 
Extensive experiments on the MeasEval benchmark and our newly constructed MeasEval-Ext dataset demonstrate that \textsc{MeasHalu} substantially reduces hallucination rates and consistently outperforms strong supervised baselines and proprietary LLMs. 
Furthermore, we show that \textsc{MeasHalu} functions as a reliable external measurement extraction tool that significantly improves performance on downstream embodied scientific tasks, validating its practical utility for trustworthy AI4Science systems.

Our contributions are summarized as follows:

We provide the first fine-grained analysis of scientific \emph{measurement hallucinations} in large language models, revealing their structural nature and identifying two fundamental sources of failure: unreliable quantitative reasoning and fragile relational grounding.
    
We propose \textsc{MeasHalu}, a unified reasoning-enhanced learning framework that systematically suppresses measurement hallucinations by integrating reasoning-aware supervised fine-tuning with targeted reinforcement learning via structured reward shaping.

 We construct a new out-of-distribution evaluation benchmark, \textsc{MeasEval-Ext}, and demonstrate through extensive experiments that \textsc{MeasHalu} substantially reduces hallucination rates and consistently outperforms strong supervised baselines and proprietary LLMs on scientific measurement extraction.

We further show that \textsc{MeasHalu} serves as a reliable external measurement extraction tool that significantly improves performance on downstream embodied scientific tasks, validating its practical utility for trustworthy AI4Science systems.

\section{Related Work}
\label{sec:related_work}

\subsection{Hallucinations in Large Language Models}
Hallucination, where language models generate ungrounded or factually incorrect content, has been extensively studied in general-purpose LLMs~\cite{huang2025survey}. 
Most prior work focuses on semantic and factual hallucinations in open-ended generation~\cite{ji2023survey}, with typical taxonomies including fabrication, inconsistency, and logical errors~\cite{li2025loki}. 
However, these taxonomies are largely developed for free-form text generation and do not capture the structural requirements of measurement extraction, where numerical faithfulness, unit consistency, and entity-quantity relational grounding are essential. 
We address this gap by proposing a fine-grained taxonomy of \emph{measurement-specific hallucinations} and designing mitigation mechanisms tailored to these failure modes.

\subsection{General Information Extraction vs.\ Measurement Extraction}
Information extraction (IE) and named entity recognition (NER) are foundational NLP tasks~\cite{nadeau2007survey}. 
While early systems relied on rule-based and feature-engineered pipelines, modern approaches increasingly leverage neural architectures and pre-trained language models. 
Nevertheless, \emph{scientific measurement extraction} poses additional constraints beyond conventional IE: models must accurately capture numerical values, units, and modifiers, and preserve their structured relations to measured entities and properties under strict grounding. 
These constraints make the task particularly sensitive to hallucinations and motivate learning objectives that explicitly penalize fabrication, mis-scoping, and relational incompleteness.

\subsection{Scientific Measurement Extraction and Benchmarks}
Scientific information extraction has been advanced by datasets such as \textsc{SciERC}~\cite{luan2018multi} and \textsc{MeasEval}~\cite{harper2021semeval}. 
Among them, \textsc{MeasEval} provides the most fine-grained annotation schema for scientific measurements, including quantities, units, modifiers, and their relations, and has become a key benchmark for evaluating measurement extraction systems. 
Despite progress, numerically grounded and relation-consistent extraction remains challenging, especially for complex sentences containing multiple measurements and implicit constraints~\cite{xu2024numcot}. 
Our work builds on the \textsc{MeasEval} schema and targets these persistent failure modes with a hallucination-aware optimization framework.

\subsection{Mitigation Strategies for LLM Hallucinations}
A wide range of techniques have been proposed to reduce hallucinations in LLMs, including retrieval-augmented generation (RAG)~\cite{lewis2020retrieval}, supervised fine-tuning (SFT)~\cite{zhou2023lima}, chain-of-thought prompting~\cite{wei2022chain}, process-based supervision~\cite{lightman2023let}, reinforcement learning from human feedback (RLHF)~\cite{ouyang2022training}, and direct preference optimization (DPO)~\cite{rafailov2023direct}. 
While effective for open-ended generation, these methods are not explicitly designed to enforce the strict grounding and structural consistency required by scientific measurement extraction. 
In contrast, our approach integrates reasoning-aware SFT with targeted reinforcement learning and structured reward shaping, explicitly encoding measurement-specific constraints to suppress hallucinations at their structural root.

Despite significant progress in hallucination mitigation, prior work has neither systematically characterized hallucinations in scientific measurement extraction nor introduced specialized reward objectives tailored to its error patterns. 
We bridge this gap by unifying a fine-grained hallucination taxonomy with a progressive optimization framework designed specifically for measurement-specific error suppression.

\section{Methodology}
\begin{figure*}[t]
    \centering
    \includegraphics[width=\textwidth]{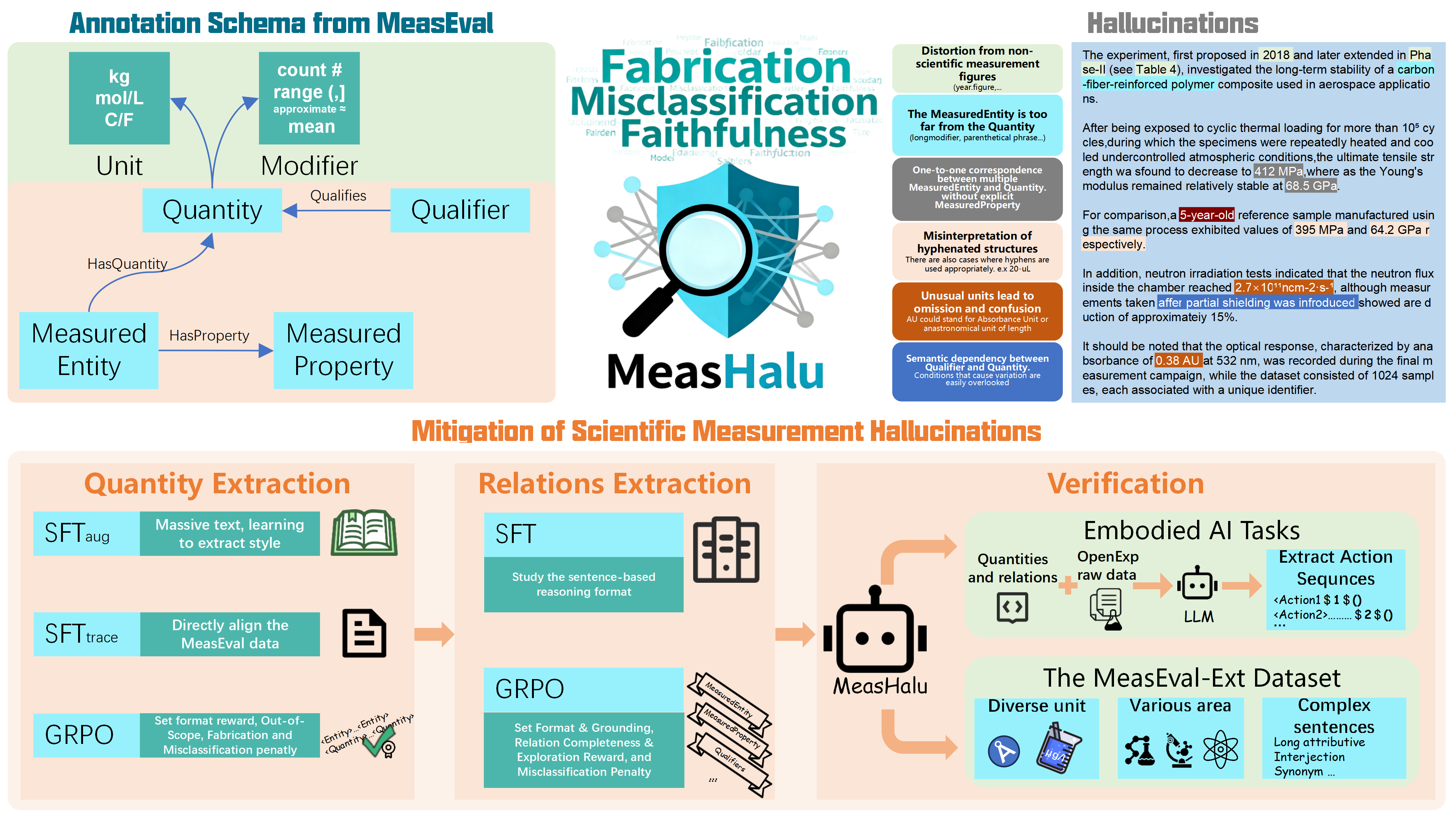}
    \caption{Overview of our method consisting of two stages, Supervised Fine-Tuning \& GRPO based Reinforcement Learning.}
    \label{fig:task}
\end{figure*}

Informed by our analysis in Section~\ref{sec:related_work}, we design \textsc{MeasHalu} around a central hypothesis: 
\emph{scientific measurement hallucinations arise from two fundamentally different failure modes—unreliable quantitative reasoning and fragile relational grounding}. 
Accordingly, our framework adopts a two-branch mitigation strategy, targeting \textbf{Quantity Hallucinations} and \textbf{Relation-based Hallucinations} respectively. 
As illustrated in Figure~\ref{fig:task}, \textsc{MeasHalu} integrates progressive supervised fine-tuning with hallucination-aware reinforcement learning, enabling the model to internalize strict scientific grounding constraints directly into its reasoning process.

These hallucinations~(see Table~\ref{tab:hallucination_taxonomy}) significantly undermine the reliability of LLMs for this critical task. 

\subsection{Quantity Hallucination Mitigation}

Unlike prior approaches that employ end-to-end joint training for quantities and relations, our method first trains quantity extraction independently. Furthermore, following the SFT stage, we incorporate a GRPO phase specifically driven by hallucination-targeted rewards to further mitigate hallucinations.

\subsubsection{Progressive Supervised Fine-tuning}
To endow the LLM with structured quantity reasoning capabilities, we adopt a progressive SFT strategy. Specifically, we first utilize $\mathcal{D}_{\text{aug}}$ to establish foundational quantity reasoning skills, followed by fine-tuning on $\mathcal{D}_{\text{trace}}$ to ensure rigorous alignment with MeasEval standards.
The construction details of these two datasets are elaborated below.
\paragraph{$\mathcal{D}_{\text{aug}}$}
We curate an unlabeled corpus $\mathcal{X}_{\text{un}}$ from arXiv paper abstracts~\citep{Cohan_2018}. Lacking gold quantity annotations, we use Quantulum3 ($f_{\text{qtm}}$, \footnote{https://github.com/nielstron/quantulum3}) to extract quantity candidates, then leverage an augmentation template $\mathcal{P}_{\text{aug}}$ to prompt $\mathcal{M}$ to verify these anchors and generate a reasoning trajectory $h_{\text{aug}}$.
Formally, for $x \in \mathcal{X}_{\text{un}}$:
\begin{align}
    \tilde{y} &\leftarrow f_{\text{qtm}}(x),\quad h_{\text{aug}} \leftarrow \mathcal{M}(x, \tilde{y}; \mathcal{P}_{\text{aug}})
\end{align}
where $\tilde{y}$ is noisy pseudo-labels from $f_{\text{qtm}}$, and $\mathcal{P}_{\text{aug}}$ guides $\mathcal{M}$ to filter false positives via semantics. Valid trajectories form $\mathcal{D}_{\text{aug}} = \{(x, h_{\text{aug}})\}_{i=1}^{20K}$.

\paragraph{$\mathcal{D}_{\text{trace}}$}
We leverage the training split of the MeasEval dataset, which contains human-annotated gold quantity labels $y_{\text{gt}}$, and adopt a \textit{traceback template} $\mathcal{P}_{\text{trace}}$ to guide reasoning reconstruction: given $y_{\text{gt}}$, the model generates a stepwise reasoning trajectory $h_{\text{trace}}$ leading to the gold conclusion, formulated as:
\begin{equation}
    h_{\text{trace}} \leftarrow \mathcal{M}(x, y_{\text{gt}}; \mathcal{P}_{\text{trace}})
\end{equation}

To ensure correctness of the reasoning trajectory, we enforce strict consistency validation via $\mathbbm{I}(\cdot)$, where $\text{concl}(\cdot)$ extracts the final quantity from $h_{\text{trace}}$. The filtered dataset is constructed as:
\begin{equation}
    \mathcal{D}_{\text{trace}} = \{ (x, h_{\text{trace}}) \mid \mathbbm{I}\big(\text{concl}(h_{\text{trace}}), y_{\text{gt}}\big) = 1 \}
\end{equation}
The prompts $\mathcal{P}_{\text{trace}}$ and $\mathcal{P}_{\text{aug}}$ are provided in Appendix~\ref{sec:prompt}.

\subsubsection{Hallucination-targeted Reward Function}

The total reward $R(y)$ is a weighted sum of four components targeting distinct quantity-related hallucinations:
\begin{equation}
R = w_1 r_{\text{fmt}} + w_2 r_{\text{scope}} + w_3 r_{\text{fab}} + w_4 r_{\text{mis}}
\end{equation}

\paragraph{Format compliance reward ($r_{\text{fmt}}$):} 
A binary reward is assigned for strict adherence to the predefined structure {\texttt{<ARABIC>}, \dots, \texttt{<CONCLUSION>}}, enforcing schema compliance and parsability of generated reasoning chains.
    \paragraph{Out-of-scope hallucination penalty ($r_{\text{scope}}$):} 
    A penalty is imposed when the model extracts out-of-scope entities—such as figure labels (e.g., ``Fig.~1'')—that do not constitute valid numerical data. This mechanism utilizes pattern recognition to identify and penalize specific noisy strings, while simultaneously penalizing any generated answers that fail to match the ground truth, ensuring that the model avoids generating arbitrary numbers that deviate from the objective definitions.

    \paragraph{Fabrication hallucination penalty ($r_{\text{fab}}$):} 
    This penalty targets invalid quantity fabrication by verifying each extracted entity against a hybrid physical parser $\mathcal{T}_{\text{parse}}$. A penalty is triggered if the extracted string fails to be parsed as a valid physical or numerical quantity, preventing the model from inventing nonsensical values.

    \paragraph{Misclassification hallucination reward ($r_{\text{mis}}$):} 
    A reward is assigned based on token-level precision to mitigate misclassification hallucinations. This mechanism imposes a penalty if the model generates excessively long spans that erroneously incorporate surrounding components, such as the MeasuredEntity, into the quantity extraction.

\noindent Detailed mathematical derivations and implementation specifics for these reward components are provided in Appendix~\ref{sec:qty_reward}.

\subsection{Relation-based Hallucination Mitigation}
The extraction of relation-based scientific measurements is particularly challenging due to long-range contextual dependencies that frequently induce hallucinations. Compared to traditional rule-based approaches that generate answers after exhaustively processing complex constraints, our approach first pinpoints the quantity-containing sentence, with subsequent reasoning anchored to this local context to extract the quantity and its relations—eliminating cross-sentence hallucination triggers. We implement this strategy via SFT for schema establishment and GRPO for hallucination-targeted alignment. Our sentence-based strategy also provides efficiency benefits by reducing redundant global reasoning. Detailed analysis is provided in Appendix~\ref{sec:efficiency}.

\subsubsection{Quantity-Guided Relation Extraction}
Given the document text $x$ and a list of candidate quantities $\mathcal{Q}{\text{in}}$, the extraction pipeline $A{\text{halu}}$ follows a two-stage chain-of-thought reasoning process.

First, the model identifies the evidence sentences $S = {s_1, \dots, s_n}$ that contain quantities in $\mathcal{Q}{\text{in}}$:
\begin{equation}
S \leftarrow A_{\text{halu}}(x, \mathcal{Q}_{\text{in}})
\end{equation}
where $S$ denotes the target sentences.

The model then performs fine-grained reasoning over $S$ to resolve quantity attributes (e.g., units and modifiers) and associate them with their corresponding measured entities or properties, yielding the final structured relations $\mathcal{R}$.

This two-stage reasoning is learned via supervised fine-tuning on rule-derived traces from MeasEval annotations.

\subsubsection{Hallucination-targeted Reward Function}

While sentence-based extraction excels at local entity identification, it often struggles to capture long-range dependency chains (e.g., \textit{MeasuredEntity}, \textit{Qualifier}), which frequently induces inference bias and leads to the under-extraction of sparse components. To mitigate these reasoning biases and suppress the resulting hallucinations, we design a composite reward function $R$ optimized via GRPO. The reward function are formulated as: 
\begin{equation}
    R = w_{1} r_{\text{fmt}} + w_{2} r_{\text{comp}} + w_{3} r_{\text{mis}}
\end{equation}
The design rationale for each reward component is detailed below:
\paragraph{Format compliance reward ($r_{\text{fmt}}$)} A composite reward is assigned to enforce strict adherence to the quantitative schema and ensure textual grounding. It imposes two constraints: first, validating the structural segmentation of reasoning sections to prevent schema collapse; second, verifying that each extracted sentence can be mapped to a valid text span in the source document.
    
\paragraph{Relational completeness reward ($r_{\text{comp}}$)}
    To mitigate inference-induced and role-definition hallucinations  stemming from broken dependency links, this reward is designed to enforce the structural integrity of the reasoning chain. The mechanism drives comprehensive exploration through a two-tier incentive structure: (1) a stepwise reward for incremental component extraction and a weighted exploration term that prioritizes harder-to-predict components to drive model exploration (2) a completeness bonus awarded only upon full recovery of the gold-standard relation group to enforce the structural integrity of the reasoning chain. 
    
\paragraph{Misclassification hallucination reward ($r_{\text{mis}}$)} A reward is assigned based on token-level precision to mitigate misclassification hallucinations. This mechanism imposes a penalty if the model generates excessively long spans that erroneously incorporate surrounding components. The detailed mathematical formulations for these reward components are provided in Appendix~\ref{sec:rel_reward}.

\section{Experiments}
In this section, We evaluate our method on quantity extraction and relation identification using the MeasEval benchmark. Additionally, we introduce \textit{MeasEval-Ext}, a specialized dataset annotated from recent literature to target novel units and complex expressions absent from the training distribution. Further analyses are conducted including entropy dynamic analysis and its utility as a functional tool within downstream embodied AI tasks.
\subsection{Effectiveness of Quantity Hallucination Mitigation Strategies}
\paragraph{Setup}
 We utilize the Quantity subset of the MeasEval dataset as our primary evaluation benchmark. To verify the scalability and robustness of our approach, we employ the \textit{Qwen2.5-Instruct} series across three different scales (0.5B, 3B, and 7B) as the base models.

\begin{table}[h]
\centering
\setlength{\tabcolsep}{2pt}
\footnotesize
\begin{tabular}{r c c c}
\toprule
\multicolumn{1}{c}{\textbf{Model Setting}} & \textbf{0.5B} & \textbf{3B} & \textbf{7B} \\
\midrule

\multicolumn{1}{l}{MeasHalu-Quant}           & \valueWithStd{0.749}{0.006} & \valueWithStd{0.812}{0.006} & \valueWithStd{0.849}{0.006} \\
w/o ($\mathcal{D}_{\text{trace}}$ + GRPO)     & \valueWithStd{0.475}{0.011} & \valueWithStd{0.481}{0.001} & \valueWithStd{0.465}{0.011} \\
w/o ($\mathcal{D}_{\text{aug}}$ + GRPO)      & \valueWithStd{0.408}{0.028} & \valueWithStd{0.346}{0.008} & \valueWithStd{0.397}{0.027} \\
w/o  GRPO & \valueWithStd{0.596}{0.008} & \valueWithStd{0.539}{0.002} & \valueWithStd{0.585}{0.018} \\
\bottomrule
\end{tabular}
\caption{Quantity extraction performance of 0.5B, 3B and 7B models (Mean $\pm$ Std).}
\label{tab:quantity-mitigation}
\end{table}

\paragraph{Results}
Our full model \textsc{MeasHalu-Quant} achieves consistent performance advantages across all model scales in Table~\ref{tab:quantity-mitigation}. Compared to the baseline without GRPO，the integration of GRPO drives SOTA results of 0.749, 0.812, and 0.849 for 0.5B, 3B, and 7B models, validating that our rule-based reward system enables stable anti-hallucination alignment even for low-capacity models.

Using only gold-standard data (w/o ($\mathcal{D}_{\text{aug}}$ + GRPO)) gives the lowest scores (e.g., 0.346 for 3B), showing models cannot capture complex multi-domain quantitative annotation rules without prior measurement extraction schema scaffolding.

The single first stage (w/o ($\mathcal{D}_{\text{trace}}$ + GRPO)) brings marginal gains from data scaling but is suboptimal, while (w/o GRPO) delivers substantial improvements (e.g., 3B score raised to 0.539). It confirms that the 1st stage initializes quantitative schema adherence, while the 2nd stage enhances generalization across diverse scientific contexts by leveraging multi-domain scientific knowledge.

\paragraph{Ablation on Reward Components}
To further understand the contribution of each reward component, we conduct fine-grained ablation studies by removing individual reward terms from the GRPO objective. The results are shown in Table~\ref{tab:reward-ablation}.

Removing either the out-of-scope penalty ($r_{\text{scope}}$) or the fabrication penalty ($r_{\text{fab}}$) leads to consistent performance degradation across all model scales, indicating that both are critical for preventing invalid or noisy extractions and suppressing hallucinated quantities. These results confirm that different reward components address complementary failure modes, and their combination is necessary to achieve robust hallucination mitigation.

\begin{table}[h]
\centering
\setlength{\tabcolsep}{2pt}
\footnotesize
\begin{tabular}{r c c c}
\toprule
\multicolumn{1}{c}{\textbf{Model Setting}} & \textbf{0.5B} & \textbf{3B} & \textbf{7B} \\
\midrule

\multicolumn{1}{l}{MeasHalu-Quant} & \valueWithStd{0.749}{0.006} & \valueWithStd{0.812}{0.006} & \valueWithStd{0.849}{0.006} \\
w/o $r_{\text{scope}}$ & \valueWithStd{0.702}{0.016} & \valueWithStd{0.735}{0.008} & \valueWithStd{0.739}{0.007} \\
w/o $r_{\text{fab}}$   & \valueWithStd{0.582}{0.011} & \valueWithStd{0.779}{0.005} & \valueWithStd{0.792}{0.006} \\

\bottomrule
\end{tabular}
\caption{Ablation study of reward components for quantity hallucination mitigation (Mean $\pm$ Std).}
\label{tab:reward-ablation}
\end{table}

\subsection{Effectiveness of Relation-based Hallucination Mitigation Strategies}
In this section, we evaluate our relation-based hallucination mitigation method on the MeasEval dataset and \textit{MeasEval-Ext}, a newly annotated dataset containing novel expressions absent from MeasEval, designed to assess the model’s generalization and robustness.
\paragraph{Setup}

We use the \textit{Qwen2.5-Instruct} models (0.5B, 3B, 7B) to assess performance across parameter scales. Although MeasEval is a high-quality benchmark, its limited size and dated sources underrepresent emerging units. To evaluate robustness under distribution shift, we introduce \textit{MeasEval-Ext}, annotated strictly following the MeasEval schema.

We employ an adversarial strategy by selecting recent literature containing novel units and complex expressions absent from the training distribution, rigorously testing model generalization beyond memorized vocabulary. Annotations followed MeasEval guidelines (see Appendix~\ref{sec:annotation_details} for agreement analysis).

\begin{table*}[t]
\centering
\setlength{\tabcolsep}{2pt} 
\footnotesize
    \begin{tabular}{l c c ccc c cc c cc c cc} 
    \toprule
    \multirow{3}{*}{\textbf{Model}} & \multirow{3}{*}{\textbf{Overall}} & & \multicolumn{3}{c}{\textbf{Quantities}} & & \multicolumn{2}{c}{\textbf{Entities}} & & \multicolumn{2}{c}{\textbf{Properties}} & & \multicolumn{2}{c}{\textbf{Qualifiers}} \\
    \cmidrule{4-6}\cmidrule{8-9}\cmidrule{11-12}\cmidrule{14-15}
     & & & \textbf{Quantity} & \textbf{Unit} & \textbf{Modifier} & & \textbf{ME} & \textbf{HasQuantity} & & \textbf{MP} & \textbf{HasProperty} & & \textbf{Qualifier} & \textbf{Qualifies} \\
    \midrule    
    \multicolumn{15}{c}{\textit{Top Ranked Systems from the MeasEval Competition}} \\ 
    \midrule
    Baseline & 0.239 & & 0.827 & 0.561 & 0.000 & & 0.053 & 0.075 & & 0.064 & 0.007 & & 0.005 & 0.000 \\
    Counts & 0.432 & & \cellcolor{lightred}{0.861} & 0.804 & \cellcolor{lightyellow}{0.614} & & 0.406 & 0.311 & & 0.245 & 0.183 & & 0.077 & 0.064 \\
    CONNER & 0.473 & & \cellcolor{lightyellow}{0.855} & 0.719 & 0.523 & & 0.398 & 0.424 & & \cellcolor{lightteal}{0.437} & 0.257 & & 0.000 & 0.000 \\
    LIORI & \cellcolor{lightred}{0.519} & & \cellcolor{lightred}{0.861} & 0.722 & \cellcolor{lightred}{0.642} & & \cellcolor{lightyellow}{0.437} & \cellcolor{lightred}{0.482} & & \cellcolor{lightred}{0.467} & \cellcolor{lightred}{0.318} & & \cellcolor{lightyellow}{0.163} & \cellcolor{lightyellow}{0.092} \\
    \midrule
    \multicolumn{15}{c}{\textit{Rule-based Prompting}} \\ 
    \midrule
    Qwen2.5-7b-inst &\valueWithoutStd{0.171}{0.028}&&\valueWithoutStd{0.491}{0.052}&\valueWithoutStd{0.478}{0.075}&\valueWithoutStd{0.106}{0.011}&&\valueWithoutStd{0.088}{0.021}&\valueWithoutStd{0.045}{0.015}&&\valueWithoutStd{0.057}{0.008}&\valueWithoutStd{0.000}{0.000}&&\valueWithoutStd{0.040}{0.012}&\valueWithoutStd{0.017}{0.006}\\    
    Qwen2.5-72b-inst &\valueWithoutStd{0.286}{0.028}&&\valueWithoutStd{0.644}{0.035}&\valueWithoutStd{0.826}{0.026}&\valueWithoutStd{0.236}{0.115}&&\valueWithoutStd{0.196}{0.038}&\valueWithoutStd{0.147}{0.028}&&\valueWithoutStd{0.164}{0.042}&\valueWithoutStd{0.001}{0.002}&&\valueWithoutStd{0.076}{0.019}&\valueWithoutStd{0.021}{0.011}\\
    DeepSeek-R1 &\valueWithoutStd{0.253}{0.008}&&\valueWithoutStd{0.569}{0.002}&\valueWithoutStd{0.586}{0.002}&\valueWithoutStd{0.240}{0.018}&&\valueWithoutStd{0.216}{0.017}&\valueWithoutStd{0.163}{0.014}&&\valueWithoutStd{0.163}{0.003}&\valueWithoutStd{0.024}{0.010}&&\valueWithoutStd{0.085}{0.015}&\valueWithoutStd{0.029}{0.002}\\
    DeepSeek-V3 &\valueWithoutStd{0.271}{0.008}&&\valueWithoutStd{0.657}{0.018}&\valueWithoutStd{0.768}{0.010}&\valueWithoutStd{0.214}{0.003}&&\valueWithoutStd{0.239}{0.009}&\valueWithoutStd{0.135}{0.014}&&\valueWithoutStd{0.113}{0.011}&\valueWithoutStd{0.001}{0.002}&&\valueWithoutStd{0.085}{0.004}&\valueWithoutStd{0.014}{0.004}\\
    Gemini-2.5-Pro &\valueWithoutStd{0.359}{0.008}&&\valueWithoutStd{0.712}{0.007}&\valueWithoutStd{0.784}{0.035}&\valueWithoutStd{0.464}{0.009}&&\valueWithoutStd{0.306}{0.009}&\valueWithoutStd{0.266}{0.007}&&\valueWithoutStd{0.287}{0.021}&\valueWithoutStd{0.090}{0.009}&&\valueWithoutStd{0.146}{0.018}&\cellcolor{lightteal}{\valueWithoutStd{0.076}{0.014}}\\
    GPT-5 &\valueWithoutStd{0.371}{0.004}&&\valueWithoutStd{0.804}{0.013}&\valueWithoutStd{0.742}{0.018}&\valueWithoutStd{0.395}{0.020}&&\valueWithoutStd{0.361}{0.003}&\valueWithoutStd{0.270}{0.005}&&\valueWithoutStd{0.355}{0.014}&\valueWithoutStd{0.020}{0.004}&&\valueWithoutStd{0.152}{0.019}&\valueWithoutStd{0.052}{0.006}\\
    \midrule
    \multicolumn{15}{c}{\textit{Sentence-based Prompting}} \\
    \midrule
    Qwen2.5-7b-inst &\valueWithoutStd{0.073}{0.006}&&\valueWithoutStd{0.151}{0.008}&\valueWithoutStd{0.160}{0.007}&\valueWithoutStd{0.027}{0.001}&&\valueWithoutStd{0.066}{0.009}&\valueWithoutStd{0.059}{0.010}&&\valueWithoutStd{0.044}{0.010}&\valueWithoutStd{0.031}{0.010}&&\valueWithoutStd{0.003}{0.004}&\valueWithoutStd{0.005}{0.006}\\    
    Qwen2.5-72b-inst &\valueWithoutStd{0.212}{0.002}&&\valueWithoutStd{0.403}{0.006}&\valueWithoutStd{0.516}{0.005}&\valueWithoutStd{0.232}{0.010}&&\valueWithoutStd{0.204}{0.005}&\valueWithoutStd{0.137}{0.007}&&\valueWithoutStd{0.113}{0.008}&\valueWithoutStd{0.087}{0.005}&&\valueWithoutStd{0.038}{0.011}&\valueWithoutStd{0.012}{0.007}\\
    DeepSeek-R1 &\valueWithoutStd{0.304}{0.004}&&\valueWithoutStd{0.589}{0.016}&\valueWithoutStd{0.711}{0.021}&\valueWithoutStd{0.356}{0.030}&&\valueWithoutStd{0.260}{0.011}&\valueWithoutStd{0.198}{0.012}&&\valueWithoutStd{0.182}{0.006}&\valueWithoutStd{0.118}{0.008}&&\valueWithoutStd{0.113}{0.030}&\valueWithoutStd{0.057}{0.019}\\  
    DeepSeek-V3 &\valueWithoutStd{0.320}{0.006}&&\valueWithoutStd{0.567}{0.013}&\valueWithoutStd{0.726}{0.005}&\valueWithoutStd{0.355}{0.016}&&\valueWithoutStd{0.303}{0.006}&\valueWithoutStd{0.225}{0.018}&&\valueWithoutStd{0.226}{0.022}&\valueWithoutStd{0.149}{0.002}&&\valueWithoutStd{0.019}{0.009}&\valueWithoutStd{0.000}{0.000}\\    
    Gemini-2.5-Pro &\valueWithoutStd{0.440}{0.011}&&\valueWithoutStd{0.782}{0.003}&\cellcolor{lightred}{\valueWithoutStd{0.882}{0.003}}&\valueWithoutStd{0.486}{0.011}&&\cellcolor{lightteal}{\valueWithoutStd{0.436}{0.017}}&\valueWithoutStd{0.376}{0.020}&&\valueWithoutStd{0.386}{0.028}&\cellcolor{lightteal}{\valueWithoutStd{0.280}{0.025}}&&\valueWithoutStd{0.143}{0.019}&\valueWithoutStd{0.056}{0.010}\\
    GPT-5 &\valueWithoutStd{0.406}{0.008}&&\valueWithoutStd{0.724}{0.007}&\valueWithoutStd{0.817}{0.017}&\valueWithoutStd{0.500}{0.027}&&\valueWithoutStd{0.397}{0.002}&\valueWithoutStd{0.351}{0.006}&&\valueWithoutStd{0.355}{0.009}&\valueWithoutStd{0.226}{0.002}&&\valueWithoutStd{0.138}{0.026}&\valueWithoutStd{0.042}{0.031}\\    
    \midrule
    MeasHalu-0.5B&\valueWithoutStd{0.347}{0.003}&&\valueWithoutStd{0.659}{0.009}&\valueWithoutStd{0.760}{0.015}&\valueWithoutStd{0.268}{0.007}&&\valueWithoutStd{0.284}{0.002}&\valueWithoutStd{0.279}{0.006}&&\valueWithoutStd{0.287}{0.006}&\valueWithoutStd{0.184}{0.003}&&\valueWithoutStd{0.059}{0.017}&\valueWithoutStd{0.037}{0.010}\\
     \multicolumn{1}{r}{w/o GRPO}  &\valueWithoutStd{0.312}{0.008}&&\valueWithoutStd{0.649}{0.001}&\valueWithoutStd{0.717}{0.019}&\valueWithoutStd{0.239}{0.006}&&\valueWithoutStd{0.263}{0.004}&\valueWithoutStd{0.241}{0.017}&&\valueWithoutStd{0.238}{0.017}&\valueWithoutStd{0.140}{0.004}&&\valueWithoutStd{0.069}{0.009}&\valueWithoutStd{0.022}{0.007}\\
    MeasHalu-3B &\valueWithoutStd{0.460}{0.005}&&\valueWithoutStd{0.810}{0.004}&\cellcolor{lightyellow}{\valueWithoutStd{0.875}{0.005}}&\valueWithoutStd{0.440}{0.002}&&\valueWithoutStd{0.396}{0.007}&\valueWithoutStd{0.418}{0.007}&&\valueWithoutStd{0.407}{0.006}&\valueWithoutStd{0.277}{0.007}&&\valueWithoutStd{0.110}{0.005}&\valueWithoutStd{0.049}{0.010}\\
     \multicolumn{1}{r}{w/o GRPO}  &\valueWithoutStd{0.433}{0.010}&&\valueWithoutStd{0.782}{0.014}&\valueWithoutStd{0.850}{0.003}&\valueWithoutStd{0.449}{0.009}&&\valueWithoutStd{0.370}{0.006}&\valueWithoutStd{0.377}{0.017}&&\valueWithoutStd{0.365}{0.019}&\valueWithoutStd{0.245}{0.002}&&\valueWithoutStd{0.084}{0.006}&\valueWithoutStd{0.043}{0.015}\\
    MeasHalu-7B &\cellcolor{lightyellow}{\valueWithoutStd{0.512}{0.004}}&&\cellcolor{lightteal}{\valueWithoutStd{0.848}{0.001}}&\valueWithoutStd{0.860}{0.008}&\valueWithoutStd{0.607}{0.006}&&\cellcolor{lightred}{\valueWithoutStd{0.455}{0.008}}&\cellcolor{lightyellow}{\valueWithoutStd{0.472}{0.009}}&&\cellcolor{lightyellow}{\valueWithoutStd{0.442}{0.012}}&\cellcolor{lightyellow}{\valueWithoutStd{0.310}{0.005}}&&\cellcolor{lightred}{\valueWithoutStd{0.170}{0.005}}&\cellcolor{lightred}{\valueWithoutStd{0.100}{0.009}}\\
     \multicolumn{1}{r}{w/o GRPO} &\cellcolor{lightteal}{\valueWithoutStd{0.479}{0.005}}&&\valueWithoutStd{0.846}{0.002}&\cellcolor{lightteal}{\valueWithoutStd{0.863}{0.004}}&\cellcolor{lightteal}{\valueWithoutStd{0.610}{0.015}}&&\valueWithoutStd{0.429}{0.004}&\cellcolor{lightteal}{\valueWithoutStd{0.433}{0.016}}&&\valueWithoutStd{0.397}{0.016}&\valueWithoutStd{0.272}{0.012}&&\cellcolor{lightteal}{\valueWithoutStd{0.155}{0.012}}&\valueWithoutStd{0.063}{0.010}\\     
    \bottomrule
    \end{tabular}
\caption{Experimental results over the MeasEval Benchmark. Comparing MeasHalu with competition leaders and rule/sentence-based LLM baselines. Top ranks are shaded orange (1st), yellow (2nd), and teal (3rd).}
\label{tab:relation-mitigation_nostd}
\end{table*}

\begin{table*}[!htp]
\centering
\setlength{\tabcolsep}{2pt} 
\footnotesize
    \begin{tabular}{l c c ccc c cc c cc c cc} 
    \toprule
    \multirow{3}{*}{\textbf{Model}} & \multirow{3}{*}{\textbf{Overall}} & & \multicolumn{3}{c}{\textbf{Quantities}} & & \multicolumn{2}{c}{\textbf{Entities}} & & \multicolumn{2}{c}{\textbf{Properties}} & & \multicolumn{2}{c}{\textbf{Qualifiers}} \\
    \cmidrule{4-6}\cmidrule{8-9}\cmidrule{11-12}\cmidrule{14-15}
     & & & \textbf{Quantity} & \textbf{Unit} & \textbf{Modifier} & & \textbf{ME} & \textbf{HasQuantity} & & \textbf{MP} & \textbf{HasProperty} & & \textbf{Qualifier} & \textbf{Qualifies} \\
    \midrule

    \multicolumn{15}{c}{\textit{Rule-based Prompting}} \\ 
    \midrule
    
    GPT-5 & \valueWithoutStd{0.383}{0.004} && \cellcolor{lightyellow}{\valueWithoutStd{0.833}{0.021}} & \valueWithoutStd{0.706}{0.017} & \valueWithoutStd{0.357}{0.031} && \valueWithoutStd{0.400}{0.004} & \valueWithoutStd{0.282}{0.003} && \cellcolor{lightteal}{\valueWithoutStd{0.310}{0.020}} & \valueWithoutStd{0.046}{0.011} && \cellcolor{lightteal}{\valueWithoutStd{0.135}{0.017}} & \valueWithoutStd{0.019}{0.012} \\
    DeepSeek-R1 & \valueWithoutStd{0.252}{0.006} && \valueWithoutStd{0.553}{0.016} & \valueWithoutStd{0.514}{0.046} & \valueWithoutStd{0.217}{0.018} && \valueWithoutStd{0.213}{0.005} & \valueWithoutStd{0.169}{0.009} && \valueWithoutStd{0.141}{0.015} & \valueWithoutStd{0.045}{0.006} && \valueWithoutStd{0.099}{0.009} & \valueWithoutStd{0.052}{0.018} \\
    DeepSeek-V3 & \valueWithoutStd{0.312}{0.003} && \valueWithoutStd{0.724}{0.014} & \valueWithoutStd{0.726}{0.011} & \valueWithoutStd{0.234}{0.009} && \valueWithoutStd{0.270}{0.014} & \valueWithoutStd{0.189}{0.003} && \valueWithoutStd{0.121}{0.005} & \valueWithoutStd{0.012}{0.003} && \valueWithoutStd{0.113}{0.018} & \valueWithoutStd{0.037}{0.006} \\
    Gemini-2.5-Pro & \valueWithoutStd{0.386}{0.009} && \valueWithoutStd{0.766}{0.005} & \valueWithoutStd{0.707}{0.012} & \valueWithoutStd{0.444}{0.026} && \valueWithoutStd{0.351}{0.014} & \cellcolor{lightteal}{\valueWithoutStd{0.312}{0.014}} && \valueWithoutStd{0.291}{0.028} & \valueWithoutStd{0.151}{0.017} && \cellcolor{lightyellow}{\valueWithoutStd{0.140}{0.017}} & \valueWithoutStd{0.055}{0.006} \\
    Qwen2.5-72b & \valueWithoutStd{0.296}{0.007} && \valueWithoutStd{0.675}{0.012} & \valueWithoutStd{0.747}{0.010} & \valueWithoutStd{0.203}{0.025} && \valueWithoutStd{0.246}{0.015} & \valueWithoutStd{0.187}{0.015} && \valueWithoutStd{0.142}{0.004} & \valueWithoutStd{0.000}{0.000} && \valueWithoutStd{0.091}{0.014} & \valueWithoutStd{0.066}{0.009} \\
    Qwen2.5-7b & \valueWithoutStd{0.181}{0.009} && \valueWithoutStd{0.501}{0.016} & \valueWithoutStd{0.353}{0.031} & \valueWithoutStd{0.148}{0.037} && \valueWithoutStd{0.108}{0.015} & \valueWithoutStd{0.078}{0.010} && \valueWithoutStd{0.069}{0.012} & \valueWithoutStd{0.001}{0.001} && \valueWithoutStd{0.038}{0.025} & \valueWithoutStd{0.006}{0.004} \\

    \midrule
    \multicolumn{15}{c}{\textit{Sentence-based Prompting}} \\ 
    \midrule
    
    GPT-5 & \cellcolor{lightteal}{\valueWithoutStd{0.402}{0.007}} && \valueWithoutStd{0.750}{0.002} & \cellcolor{lightteal}{\valueWithoutStd{0.759}{0.005}} & \cellcolor{lightyellow}{\valueWithoutStd{0.458}{0.007}} && \cellcolor{lightteal}{\valueWithoutStd{0.435}{0.013}} & \valueWithoutStd{0.303}{0.007} && \valueWithoutStd{0.303}{0.010} & \cellcolor{lightteal}{\valueWithoutStd{0.239}{0.018}} && \valueWithoutStd{0.100}{0.016} & \cellcolor{lightteal}{\valueWithoutStd{0.070}{0.011}} \\
    DeepSeek-R1 & \valueWithoutStd{0.324}{0.013} && \valueWithoutStd{0.622}{0.021} & \valueWithoutStd{0.648}{0.018} & \valueWithoutStd{0.255}{0.012} && \valueWithoutStd{0.299}{0.014} & \valueWithoutStd{0.235}{0.019} && \valueWithoutStd{0.215}{0.008} & \valueWithoutStd{0.146}{0.013} && \valueWithoutStd{0.075}{0.015} & \valueWithoutStd{0.069}{0.014} \\
    DeepSeek-V3 & \valueWithoutStd{0.299}{0.012} && \valueWithoutStd{0.521}{0.023} & \valueWithoutStd{0.565}{0.020} & \valueWithoutStd{0.229}{0.007} && \valueWithoutStd{0.296}{0.017} & \valueWithoutStd{0.229}{0.012} && \valueWithoutStd{0.209}{0.016} & \valueWithoutStd{0.155}{0.008} && \valueWithoutStd{0.053}{0.004} & \valueWithoutStd{0.048}{0.004} \\
    Gemini-2.5-Pro & \cellcolor{lightyellow}{\valueWithoutStd{0.462}{0.007}} && \cellcolor{lightteal}{\valueWithoutStd{0.827}{0.009}} & \cellcolor{lightred}{\valueWithoutStd{0.832}{0.007}} & \cellcolor{lightteal}{\valueWithoutStd{0.444}{0.010}} && \cellcolor{lightyellow}{\valueWithoutStd{0.472}{0.014}} & \cellcolor{lightyellow}{\valueWithoutStd{0.399}{0.012}} && \cellcolor{lightyellow}{\valueWithoutStd{0.402}{0.006}} & \cellcolor{lightyellow}{\valueWithoutStd{0.332}{0.015}} && \valueWithoutStd{0.100}{0.002} & \cellcolor{lightyellow}{\valueWithoutStd{0.072}{0.003}} \\
    Qwen2.5-72b & \valueWithoutStd{0.202}{0.006} && \valueWithoutStd{0.343}{0.008} & \valueWithoutStd{0.383}{0.006} & \valueWithoutStd{0.183}{0.007} && \valueWithoutStd{0.207}{0.012} & \valueWithoutStd{0.145}{0.012} && \valueWithoutStd{0.127}{0.007} & \valueWithoutStd{0.106}{0.010} && \valueWithoutStd{0.048}{0.001} & \valueWithoutStd{0.050}{0.007} \\
    Qwen2.5-7b & \valueWithoutStd{0.033}{0.003} && \valueWithoutStd{0.065}{0.002} & \valueWithoutStd{0.056}{0.005} & \valueWithoutStd{0.022}{0.005} && \valueWithoutStd{0.030}{0.001} & \valueWithoutStd{0.028}{0.006} && \valueWithoutStd{0.017}{0.005} & \valueWithoutStd{0.008}{0.003} && \valueWithoutStd{0.016}{0.007} & \valueWithoutStd{0.020}{0.007} \\
        
    \midrule
    MeasHalu-7B & \cellcolor{lightred}{\valueWithoutStd{0.578}{0.002}} && \cellcolor{lightred}{\valueWithoutStd{0.861}{0.003}} & \cellcolor{lightred}{\valueWithoutStd{0.832}{0.012}} & \cellcolor{lightred}{\valueWithoutStd{0.539}{0.005}} && \cellcolor{lightred}{\valueWithoutStd{0.555}{0.006}} & \cellcolor{lightred}{\valueWithoutStd{0.551}{0.008}} && \cellcolor{lightred}{\valueWithoutStd{0.522}{0.006}} & \cellcolor{lightred}{\valueWithoutStd{0.459}{0.007}} && \cellcolor{lightred}{\valueWithoutStd{0.159}{0.021}} & \cellcolor{lightred}{\valueWithoutStd{0.097}{0.004}} \\

    \bottomrule
    \end{tabular}
\caption{Experimental results over the MeasEval-Ext.}
\label{tab:results_self_dataset}
\end{table*}

\paragraph{Results over MeasEval}

Table~\ref{tab:relation-mitigation_nostd} compares complex quantitative relation extraction on the MeasEval test set. \textsc{MeasHalu-7B} achieves an overall F1 of \textbf{0.512}, closely matching the competition winner \textit{LIORI}~\citep{davletov-etal-2021-liori-semeval} (0.519)\footnote{LIORI uses a six-model ensemble and does not release weights.}, and substantially outperforming other supervised baselines such as \textit{CONNER}~\citep{cao-etal-2021-conner} (0.473) and \textit{Counts}~\citep{gangwar-etal-2021-counts} (0.432).

\begin{table*}[!htp]
\centering
\setlength{\tabcolsep}{1pt} 
\scriptsize
\resizebox{\linewidth}{!}{
\begin{tabular}{l c c ccc c cc c cc c cc} 
\toprule
\multirow{3}{*}{\textbf{Model Setting}} & \multirow{3}{*}{\textbf{Overall}} & & \multicolumn{3}{c}{\textbf{Quantities}} & & \multicolumn{2}{c}{\textbf{Entities}} & & \multicolumn{2}{c}{\textbf{Properties}} & & \multicolumn{2}{c}{\textbf{Qualifiers}} \\
\cmidrule{4-6}\cmidrule{8-9}\cmidrule{11-12}\cmidrule{14-15}
 & & & \textbf{Quantity} & \textbf{Unit} & \textbf{Modifier} & & \textbf{ME} & \textbf{HasQuantity} & & \textbf{MP} & \textbf{HasProperty} & & \textbf{Qualifier} & \textbf{Qualifies} \\
\midrule

\multicolumn{15}{c}{\textit{7B Model}} \\
\midrule
MeasHalu-7B & \valueWithStd{0.512}{0.004} && \valueWithStd{0.848}{0.001} & \valueWithStd{0.860}{0.008} & \valueWithStd{0.607}{0.006} && \valueWithStd{0.455}{0.008} & \valueWithStd{0.472}{0.009} && \valueWithStd{0.442}{0.012} & \valueWithStd{0.310}{0.005} && \valueWithStd{0.170}{0.005} & \valueWithStd{0.100}{0.009} \\
\multicolumn{1}{r}{w/o $r_{\text{mis}}$}  & \valueWithStd{0.481}{0.001} && \valueWithStd{0.852}{0.004} & \valueWithStd{0.867}{0.009} & \valueWithStd{0.599}{0.012} && \valueWithStd{0.445}{0.007} & \valueWithStd{0.443}{0.005} && \valueWithStd{0.397}{0.004} & \valueWithStd{0.277}{0.006} && \valueWithStd{0.141}{0.006} & \valueWithStd{0.041}{0.003} \\
\multicolumn{1}{r}{w/o $r_{\text{comp}}$} & \valueWithStd{0.485}{0.005} && \valueWithStd{0.845}{0.003} & \valueWithStd{0.857}{0.011} & \valueWithStd{0.606}{0.013} && \valueWithStd{0.433}{0.019} & \valueWithStd{0.406}{0.008} && \valueWithStd{0.389}{0.005} & \valueWithStd{0.281}{0.009} && \valueWithStd{0.126}{0.013} & \valueWithStd{0.070}{0.002} \\

\midrule
\multicolumn{15}{c}{\textit{3B Model}} \\
\midrule
MeasHalu-3B & \valueWithStd{0.460}{0.005} && \valueWithStd{0.810}{0.004} & \valueWithStd{0.875}{0.005} & \valueWithStd{0.440}{0.002} && \valueWithStd{0.396}{0.007} & \valueWithStd{0.418}{0.007} && \valueWithStd{0.407}{0.006} & \valueWithStd{0.277}{0.007} && \valueWithStd{0.110}{0.005} & \valueWithStd{0.049}{0.010} \\
\multicolumn{1}{r}{w/o $r_{\text{mis}}$}  & \valueWithStd{0.451}{0.006} && \valueWithStd{0.815}{0.004} & \valueWithStd{0.874}{0.002} & \valueWithStd{0.450}{0.006} && \valueWithStd{0.398}{0.011} & \valueWithStd{0.412}{0.005} && \valueWithStd{0.400}{0.006} & \valueWithStd{0.274}{0.012} && \valueWithStd{0.092}{0.018} & \valueWithStd{0.033}{0.006} \\
\multicolumn{1}{r}{w/o $r_{\text{comp}}$} & \valueWithStd{0.451}{0.002} && \valueWithStd{0.819}{0.004} & \valueWithStd{0.878}{0.004} & \valueWithStd{0.472}{0.015} && \valueWithStd{0.379}{0.005} & \valueWithStd{0.405}{0.009} && \valueWithStd{0.396}{0.008} & \valueWithStd{0.257}{0.004} && \valueWithStd{0.079}{0.003} & \valueWithStd{0.030}{0.011} \\

\midrule
\multicolumn{15}{c}{\textit{0.5B Model}} \\
\midrule
MeasHalu-0.5B & \valueWithStd{0.347}{0.003} && \valueWithStd{0.659}{0.009} & \valueWithStd{0.760}{0.015} & \valueWithStd{0.268}{0.007} && \valueWithStd{0.284}{0.002} & \valueWithStd{0.279}{0.006} && \valueWithStd{0.287}{0.006} & \valueWithStd{0.184}{0.003} && \valueWithStd{0.059}{0.017} & \valueWithStd{0.037}{0.010} \\
\multicolumn{1}{r}{w/o $r_{\text{mis}}$}  & \valueWithStd{0.329}{0.006} && \valueWithStd{0.661}{0.003} & \valueWithStd{0.722}{0.004} & \valueWithStd{0.278}{0.013} && \valueWithStd{0.360}{0.114} & \valueWithStd{0.254}{0.014} && \valueWithStd{0.253}{0.015} & \valueWithStd{0.165}{0.014} && \valueWithStd{0.091}{0.004} & \valueWithStd{0.050}{0.013} \\
\multicolumn{1}{r}{w/o $r_{\text{comp}}$} & \valueWithStd{0.338}{0.003} && \valueWithStd{0.642}{0.011} & \valueWithStd{0.751}{0.016} & \valueWithStd{0.271}{0.013} && \valueWithStd{0.291}{0.006} & \valueWithStd{0.266}{0.008} && \valueWithStd{0.283}{0.008} & \valueWithStd{0.180}{0.010} && \valueWithStd{0.309}{0.377} & \valueWithStd{0.039}{0.011} \\

\bottomrule
\end{tabular}
}
\caption{Ablation study of reward components for relation hallucination mitigation across different model scales (Mean $\pm$ Std).}
\label{tab:relation-reward-ablation}
\end{table*}
Our model also surpasses state-of-the-art proprietary LLMs (e.g., GPT-5, Gemini-2.5-Pro). Even with optimized sentence-based prompting, GPT-5 reaches only 0.406 F1, leaving \textsc{MeasHalu-7B} ahead by over 10 points. This result highlights the necessity of our quantitative domain alignment pipeline (SFT + composite reward optimization) for mitigating relational Quantity hallucinations.

Across all baseline LLMs, sentence-based prompting consistently outperforms rule-based prompting (e.g., Gemini-2.5-Pro improves from 0.359 to 0.440), supporting our hypothesis that sentence-level localized reasoning is more effective than rigid global rule-based deduction for complex quantitative relation extraction.

As shown in Table~\ref{tab:results_self_dataset}, the results on \textit{MeasEval-Ext} expose a significant performance gap: while general LLMs exhibit non-uniform shifts—often struggling with novel expressions—\textsc{MeasHalu} demonstrates robust generalization to unseen distributions, substantially widening its lead over all baselines.
Detailed statistics with standard deviations can be found in Table~\ref{tab:relation-mitigation} and Table~\ref{tab:results_self_dataset_with_std}.

\paragraph{Ablation on Reward Components}
To further investigate the contribution of each reward component for relation-based hallucination mitigation, we conduct fine-grained ablation studies by removing individual reward terms from the GRPO objective. The results are presented in Table~\ref{tab:relation-reward-ablation}.

Across all model scales (0.5B, 3B, and 7B), removing any reward component consistently leads to performance degradation, indicating that each term plays an essential role in maintaining precise span boundaries, preventing over-generation, and preserving relational consistency, especially for sparse components such as \textit{Qualifier} and \textit{Qualifies}.

Overall, the full \textsc{MeasHalu} models achieve the best performance across all metrics, while ablated variants exhibit reduced robustness in either quantity prediction or relational consistency. These results confirm that the reward components address complementary aspects of relation hallucination and are jointly necessary for stable extraction. Additional experiments on cross-domain generalization are provided in Appendix~\ref{sec:generalization}.

\subsection{Further Analysis}

\paragraph{Mechanism of Hallucination Suppression via Entropy Dynamics}
\label{sec:entropy_analysis}

Inspired by ~\citet{cui2025entropy}, we quantify Cognitive Hesitation via entropy dynamics, adapting the analysis to our task by distinguishing the quantity group (Quantity, Unit, Modifier) from the relation group (MeasuredEntity, MeasuredProperty, qualifier).

We focus on tokens strictly bounded by square brackets (e.g., parsing \texttt{70 m} from the tagged sequence \texttt{...surface form [70 m]...}). 
To capture micro-level certainty, we report four key statistics: Bracket Entropy Mean ($H_B$) and Std ($\sigma_B$) measure the average confidence level; Spike Rate ($R_B$) for the proportion of brackets containing high-entropy tokens; and Sample Spike Ratio ($R_{sample}$) quantifies the proportion of samples containing at least one high-risk fluctuation.

\begin{table}[htp]
    \centering
    \footnotesize
    \setlength{\tabcolsep}{2pt}
    \begin{tabular}{llcccc}
        \toprule
        \textbf{Group} & \textbf{Metric} & \textbf{w/o GRPO} & \textbf{MeasHalu} & \textbf{Change} \\
        \midrule
        \multirow{4}{*}{Quantity} 
        & $H_B$ & 0.0071 bits & \textbf{0.0034 bits} & $\downarrow$ 52.1\% \\
        & $\sigma_B$ & 0.0729 & \textbf{0.0477} & $\downarrow$ 34.6\%  \\
        & $R_B$ & 0.16\% & 0.32\% & -  \\
        & $R_{sample}$ & 0.77\% & 1.54\% & -  \\
        \midrule
        \multirow{4}{*}{Relation} 
        & $H_B$ & 0.1147 bits & \textbf{0.0662 bits} & $\downarrow$ 42.3\%  \\
        & $\sigma_B$ & 0.3253 & \textbf{0.2359} & $\downarrow$ 27.5\% \\
        & $R_B$ & 13.01\% & \textbf{5.23\%} & $\downarrow$ 59.8\% \\
        & $R_{sample}$ & 33.85\% & \textbf{14.62\%} & $\downarrow$ 56.8\% \\
        \bottomrule
    \end{tabular}
        \caption{Fine-grained entropy statistics by semantic role.}
\label{tab:entropy_stats_grouped}
\end{table}

Table~\ref{tab:entropy_stats_grouped} reveals a clear dichotomy in hallucination suppression across semantic roles. \textit{1) Quantity Group Stability:} SFT is already near-deterministic ($H \approx 0.0071$). GRPO further compresses residual uncertainty ($H \approx 0.0034$) with negligible spike fluctuations ($R_{sample} \approx 1.54\%$). \textit{2) Relation Group Sharpening:} GRPO reduces the spike ratio from 33.85\% to 14.62\% and lowers mean entropy by 42.3\%. These results indicate that relational reasoning, which is highly ambiguous under SFT, becomes substantially more stable under GRPO. We attribute this improvement to the directed collapse induced by GRPO, which truncates long-tail uncertainty and enforces convergence toward deterministic facts.

Subsequently, to illustrate the effect of GRPO training on reasoning stability more intuitively, we select high-entropy points in the reasoning process for a case study. Details are in Appendix~\ref{app:case_study}.

\paragraph{Application for Embodied AI Tasks}
To validate the practical value of our fine-grained extraction for embodied AI, we adapt OpenExp~\cite{liu-etal-2024-reactxt} to a text-to-action generation task, where models generate executable chemical action sequences (e.g., \textit{ADD \dots (100 mg)}) from unstructured experimental text, mimicking real-world automated laboratory scenarios.

We construct \textit{OpenExp-Action-100}, a dataset of 100 diverse instances, by using unstructured experimental narratives as inputs and OpenExp’s linearized action sequences as gold-standard outputs. To enable controlled comparison, we further define three experimental settings: Baseline (no augmentation), Gemini-Aug (quantity relations extracted by Gemini) and MeasEval-Aug (quantity relations extracted by MeasHalu).

\begin{table}[htp]
\centering
\setlength{\tabcolsep}{2pt}
\scriptsize
    \begin{tabular}{l l c c c c c c c} 
  \toprule
    \multirow{2.5}{*}{\textbf{Model}} & \multirow{2.5}{*}{\textbf{Source}} & \multirow{2.5}{*}{\textbf{Val}} & \multicolumn{2}{c}{\textbf{BLEU}} & \textbf{LEV} & \multicolumn{3}{c}{\textbf{ROUGE}} \\
    \cmidrule{4-5} \cmidrule{7-9}
     & & & \textbf{B-2} & \textbf{B-4} & \textbf{50\%} & \textbf{R-1} & \textbf{R-2} & \textbf{R-L} \\
    \midrule

    \multirow{3}{*}{Gemini-2.5-Pro} 
      & MeasHalu & \textbf{\valueWithoutStd{14.67}{2.31}} & \textbf{\valueWithoutStd{58.72}{0.23}} & \textbf{\valueWithoutStd{44.23}{0.29}} & \textbf{\valueWithoutStd{60.00}{2.00}} & \textbf{\valueWithoutStd{71.71}{0.18}} & \textbf{\valueWithoutStd{54.24}{0.32}} & \textbf{\valueWithoutStd{66.51}{0.26}} \\
      & Gemini   & \valueWithoutStd{12.67}{1.53} & \valueWithoutStd{58.26}{0.60} & \valueWithoutStd{43.99}{0.67} & \valueWithoutStd{59.67}{3.06} & \valueWithoutStd{71.28}{0.39} & \valueWithoutStd{54.21}{0.55} & \valueWithoutStd{66.49}{0.44} \\
      & None     & \valueWithoutStd{0.33}{0.58}  & \valueWithoutStd{51.78}{0.32} & \valueWithoutStd{37.23}{0.36} & \valueWithoutStd{37.33}{6.66} & \valueWithoutStd{62.92}{0.39} & \valueWithoutStd{46.25}{0.27} & \valueWithoutStd{58.90}{0.26} \\
    \midrule
    
    \multirow{3}{*}{DeepSeek-R1} 
      & MeasHalu & \textbf{\valueWithoutStd{10.67}{1.53}} & \textbf{\valueWithoutStd{58.99}{0.37}} & \textbf{\valueWithoutStd{43.01}{0.56}} & \textbf{\valueWithoutStd{61.33}{2.89}} & \textbf{\valueWithoutStd{71.62}{0.19}} & \textbf{\valueWithoutStd{52.79}{0.12}} & \textbf{\valueWithoutStd{65.85}{0.40}} \\
      & Gemini   & \valueWithoutStd{10.33}{0.58} & \valueWithoutStd{58.28}{0.73} & \valueWithoutStd{42.37}{0.61} & \valueWithoutStd{56.67}{2.89} & \valueWithoutStd{71.13}{0.38} & \valueWithoutStd{52.31}{0.36} & \valueWithoutStd{65.65}{0.31} \\
      & None     & \valueWithoutStd{8.67}{1.15}  & \valueWithoutStd{38.32}{1.02} & \valueWithoutStd{26.21}{0.99} & \valueWithoutStd{22.00}{5.29} & \valueWithoutStd{59.12}{0.31} & \valueWithoutStd{42.32}{0.21} & \valueWithoutStd{53.56}{0.41} \\
        \midrule

    \multirow{3}{*}{GPT-5} 
      & MeasHalu & \textbf{\valueWithoutStd{16.33}{2.52}} & \textbf{\valueWithoutStd{51.55}{0.29}} & \valueWithoutStd{37.43}{0.52} & \textbf{\valueWithoutStd{52.33}{3.51}} & \textbf{\valueWithoutStd{71.00}{0.39}} & \valueWithoutStd{51.62}{0.27} & \textbf{\valueWithoutStd{65.47}{0.32}} \\
      & Gemini   & \valueWithoutStd{13.33}{2.08} & \valueWithoutStd{50.61}{0.72} & \valueWithoutStd{36.56}{0.81} & \valueWithoutStd{48.33}{4.04} & \valueWithoutStd{70.73}{0.80} & \valueWithoutStd{51.31}{0.64} & \valueWithoutStd{65.06}{0.66} \\
      & None     & \valueWithoutStd{1.35}{1.54}  & \valueWithoutStd{50.39}{0.58} & \textbf{\valueWithoutStd{39.39}{0.78}} & \valueWithoutStd{50.17}{2.33} & \valueWithoutStd{66.31}{0.35} & \textbf{\valueWithoutStd{52.53}{0.40}} & \valueWithoutStd{62.76}{0.44} \\
    
    \bottomrule
    \end{tabular}
\caption{Performance on OpenExp-Action-100 with MeasEval-formatted quantity–relation context from different sources (MeasHalu vs. Gemini). Best scores per model are in bold.}

\label{tab:reactxt_results_without_std}
\end{table}

Table~\ref{tab:reactxt_results_without_std} shows that injecting structured quantity relations significantly improves Structural Validity (Val, executable/logical consistency), with MeasEval-Aug (82.3\%) outperforming Gemini-Aug and Baseline. The modest BLEU improvement (19.8 vs. 16.3) stems from gold-standard granularity mismatch. Specifically, OpenExp’s minimalist annotations omit critical details (e.g., \textit{anhydrous}) that our extraction retains. For embodied AI, structural validity—rather than textual overlap—is pivotal; MeasEval-formatted extraction ensures this validity by capturing critical details, providing constraints for executable instructions and practical utility for perception-to-action pipelines. Table~\ref{tab:reactxt_results} in the Appendix shows the full table with standard deviations.

\section{Conclusion}

The proposal of \textsc{MeasHalu} marks a significant step forward in systematically characterizing measurement hallucinations in large language models within the scientific extraction domain. Experimental results on both in-distribution and newly annotated out-of-distribution benchmarks (\textit{MeasEval-Ext}) show that \textsc{MeasHalu} substantially improves robustness and consistently outperforms strong supervised baselines and state-of-the-art large language models.
 Ultimately, \textsc{MeasHalu} proves to be a reliable external tool that drives significant gains in downstream applications, validating its utility for embodied AI and AI4Science.

\section*{Limitations}

Despite advances achieved in this paper, MeasHalu has notable limitations.  First, even though MeasHalu outperforms all existing baselines, the extraction performance for sparse components (e.g., qualifiers, F1 = 0.170) remains low, hindered by limited annotated data and ambiguous semantic dependencies in scientific text. Second, the framework’s generalization to low-resource languages or domain-specific jargon (e.g., niche engineering units) is untested, as current training data focuses on English scientific literature. Third, processing ultra-long documents with nested measurement relations may introduce computational inefficiencies, as the sentence-based reasoning strategy requires contextual localization for each quantity.

\section*{Acknowledgments}
This work was partially supported by the National Natural Science Foundation of China (62406314), the China Postdoctoral Science Foundation (2023M733654), the Guangdong Basic and Applied Basic
Research Foundation (2023A1515110496).

\bibliography{custom}

@InProceedings{HyperPIE,
author="Saier, Tarek
and Ohta, Mayumi
and Asakura, Takuto
and F{\"a}rber, Michael",
editor="Goharian, Nazli
and Tonellotto, Nicola
and He, Yulan
and Lipani, Aldo
and McDonald, Graham
and Macdonald, Craig
and Ounis, Iadh",
title="HyperPIE: Hyperparameter Information Extraction from Scientific Publications",
booktitle="Advances in Information Retrieval",
year="2024",
publisher="Springer Nature Switzerland",
address="Cham",
pages="254--269",
isbn="978-3-031-56060-6"
}

@article{foppiano2024mining,
  title={Mining experimental data from materials science literature with large language models: an evaluation study},
  author={Foppiano, Luca and Lambard, Guillaume and Amagasa, Toshiyuki and Ishii, Masashi},
  journal={Science and Technology of Advanced Materials: Methods},
  volume={4},
  number={1},
  pages={2356506},
  year={2024},
  publisher={Taylor \& Francis}
}

@inproceedings{berrahou2013extract,
  title={How to extract unit of measure in scientific documents?},
  author={Berrahou, Soumia Lilia and Buche, Patrice and Dibie-Barthelemy, Juliette and Roche, Mathieu},
  booktitle={Special Session on Text Mining},
  volume={2},
  pages={249--256},
  year={2013},
  organization={SCITEPRESS}
}

@Article{Kononova2021,
author={Kononova, Olga
and He, Tanjin
and Huo, Haoyan
and Trewartha, Amalie
and Olivetti, Elsa A.
and Ceder, Gerbrand},
title={Opportunities and challenges of text mining in materials research},
journal={iScience},
year={2021},
month={Mar},
day={19},
publisher={Elsevier},
volume={24},
number={3},
issn={2589-0042},
doi={10.1016/j.isci.2021.102155},
url={https://doi.org/10.1016/j.isci.2021.102155}
}

@article{Cohan_2018,
   title={A Discourse-Aware Attention Model for Abstractive Summarization of
            Long Documents},
   url={http://dx.doi.org/10.18653/v1/n18-2097},
   DOI={10.18653/v1/n18-2097},
   journal={Proceedings of the 2018 Conference of the North American Chapter of
          the Association for Computational Linguistics: Human Language
          Technologies, Volume 2 (Short Papers)},
   publisher={Association for Computational Linguistics},
   author={Cohan, Arman and Dernoncourt, Franck and Kim, Doo Soon and Bui, Trung and Kim, Seokhwan and Chang, Walter and Goharian, Nazli},
   year={2018}
}

@article{10.1162/qss_a_00327,
    author = {Hanson, Mark A. and Barreiro, Pablo Gómez and Crosetto, Paolo and Brockington, Dan},
    title = {The strain on scientific publishing},
    journal = {Quantitative Science Studies},
    volume = {5},
    number = {4},
    pages = {823-843},
    year = {2024},
    month = {11},
    issn = {2641-3337},
    doi = {10.1162/qss_a_00327},
    url = {https://doi.org/10.1162/qss_a_00327},
    eprint = {https://direct.mit.edu/qss/article-pdf/5/4/823/2478590/qss_a_00327.pdf},
}

@misc{chen2025ai4researchsurveyartificialintelligence,
      title={AI4Research: A Survey of Artificial Intelligence for Scientific Research}, 
      author={Qiguang Chen and Mingda Yang and Libo Qin and Jinhao Liu and Zheng Yan and Jiannan Guan and Dengyun Peng and Yiyan Ji and Hanjing Li and Mengkang Hu and Yimeng Zhang and Yihao Liang and Yuhang Zhou and Jiaqi Wang and Zhi Chen and Wanxiang Che},
      year={2025},
      eprint={2507.01903},
      archivePrefix={arXiv},
      primaryClass={cs.CL},
      url={https://arxiv.org/abs/2507.01903}, 
}

@article{cui2025entropy,
  title={The Entropy Mechanism of Reinforcement Learning for Reasoning Language Models},
  author={Cui, Ganqu and Ding, Ning},
  journal={Computing Magazine of the CCF},
  volume={1},
  number={7},
  pages={26--33},
  year={2025}
}

@article{lewis2020retrieval,
  title={Retrieval-augmented generation for knowledge-intensive nlp tasks},
  author={Lewis, Patrick and Perez, Ethan and Piktus, Aleksandra and Petroni, Fabio and Karpukhin, Vladimir and Goyal, Naman and K{\"u}ttler, Heinrich and Lewis, Mike and Yih, Wen-tau and Rockt{\"a}schel, Tim and others},
  journal={Advances in neural information processing systems},
  volume={33},
  pages={9459--9474},
  year={2020}
}

@article{polak2024extracting,
  title={Extracting accurate materials data from research papers with conversational language models and prompt engineering},
  author={Polak, Maciej P and Morgan, Dane},
  journal={Nature Communications},
  volume={15},
  number={1},
  pages={1569},
  year={2024},
  publisher={Nature Publishing Group UK London}
}

@article{dagdelen2024structured,
  title={Structured information extraction from scientific text with large language models},
  author={Dagdelen, John and Dunn, Alexander and Lee, Sanghoon and Walker, Nicholas and Rosen, Andrew S and Ceder, Gerbrand and Persson, Kristin A and Jain, Anubhav},
  journal={Nature communications},
  volume={15},
  number={1},
  pages={1418},
  year={2024},
  publisher={Nature Publishing Group UK London}
}

@inproceedings{almasian2023cqe,
  title={CQE: a comprehensive quantity extractor},
  author={Almasian, Satya and Kazakova, Vivian and G{\"o}ldner, Philipp and Gertz, Michael},
  booktitle={Proceedings of the 2023 Conference on Empirical Methods in Natural Language Processing},
  pages={12845--12859},
  year={2023}
}

@article{nadeau2007survey,
  title={A survey of named entity recognition and classification},
  author={Nadeau, David and Sekine, Satoshi},
  journal={Lingvisticae Investigationes},
  volume={30},
  number={1},
  pages={3--26},
  year={2007},
  publisher={John Benjamins}
}

@article{ji2023survey,
  title={Survey of hallucination in natural language generation},
  author={Ji, Ziwei and Lee, Nayeon and Frieske, Rita and Yu, Tiezheng and Su, Dan and Xu, Yan and Ishii, Etsuko and Bang, Ye Jin and Madotto, Andrea and Fung, Pascale},
  journal={ACM computing surveys},
  volume={55},
  number={12},
  pages={1--38},
  year={2023},
  publisher={ACM New York, NY}
}

@article{li2025loki,
  title={Loki's Dance of Illusions: A Comprehensive Survey of Hallucination in Large Language Models},
  author={Li, Chaozhuo and Wang, Pengbo and Wang, Chenxu and Zhang, Litian and Liu, Zheng and Ye, Qiwei and Xu, Yuanbo and Huang, Feiran and Zhang, Xi and Yu, Philip S},
  journal={arXiv preprint arXiv:2507.02870},
  year={2025}
}

@article{huang2025survey,
  title={A survey on hallucination in large language models: Principles, taxonomy, challenges, and open questions},
  author={Huang, Lei and Yu, Weijiang and Ma, Weitao and Zhong, Weihong and Feng, Zhangyin and Wang, Haotian and Chen, Qianglong and Peng, Weihua and Feng, Xiaocheng and Qin, Bing and others},
  journal={ACM Transactions on Information Systems},
  volume={43},
  number={2},
  pages={1--55},
  year={2025},
  publisher={ACM New York, NY}
}

@inproceedings{harper2021semeval,
  title={SemEval-2021 task 8: MeasEval--extracting counts and measurements and their related contexts},
  author={Harper, Corey and Cox, Jessica and Kohler, Curt and Scerri, Antony and Daniel Jr, Ron and Groth, Paul},
  booktitle={Proceedings of the 15th International Workshop on Semantic Evaluation (SemEval-2021)},
  pages={306--316},
  year={2021}
}

@article{luan2018multi,
  title={Multi-task identification of entities, relations, and coreference for scientific knowledge graph construction},
  author={Luan, Yi and He, Luheng and Ostendorf, Mari and Hajishirzi, Hannaneh},
  journal={arXiv preprint arXiv:1808.09602},
  year={2018}
}

@article{wei2022chain,
  title={Chain-of-thought prompting elicits reasoning in large language models},
  author={Wei, Jason and Wang, Xuezhi and Schuurmans, Dale and Bosma, Maarten and Xia, Fei and Chi, Ed and Le, Quoc V and Zhou, Denny and others},
  journal={Advances in neural information processing systems},
  volume={35},
  pages={24824--24837},
  year={2022}
}

@article{zhou2023lima,
  title={Lima: Less is more for alignment},
  author={Zhou, Chunting and Liu, Pengfei and Xu, Puxin and Iyer, Srinivasan and Sun, Jiao and Mao, Yuning and Ma, Xuezhe and Efrat, Avia and Yu, Ping and Yu, Lili and others},
  journal={Advances in Neural Information Processing Systems},
  volume={36},
  pages={55006--55021},
  year={2023}
}

@article{ouyang2022training,
  title={Training language models to follow instructions with human feedback},
  author={Ouyang, Long and Wu, Jeffrey and Jiang, Xu and Almeida, Diogo and Wainwright, Carroll and Mishkin, Pamela and Zhang, Chong and Agarwal, Sandhini and Slama, Katarina and Ray, Alex and others},
  journal={Advances in neural information processing systems},
  volume={35},
  pages={27730--27744},
  year={2022}
}

@article{rafailov2023direct,
  title={Direct preference optimization: Your language model is secretly a reward model},
  author={Rafailov, Rafael and Sharma, Archit and Mitchell, Eric and Manning, Christopher D and Ermon, Stefano and Finn, Chelsea},
  journal={Advances in neural information processing systems},
  volume={36},
  pages={53728--53741},
  year={2023}
}

@inproceedings{lightman2023let,
  title={Let's verify step by step},
  author={Lightman, Hunter and Kosaraju, Vineet and Burda, Yuri and Edwards, Harrison and Baker, Bowen and Lee, Teddy and Leike, Jan and Schulman, John and Sutskever, Ilya and Cobbe, Karl},
  booktitle={The Twelfth International Conference on Learning Representations},
  year={2023}
}

@inproceedings{xu2024numcot,
  title={NUMCoT: Numerals and units of measurement in chain-of-thought reasoning using large language models},
  author={Xu, Ancheng and Tan, Minghuan and Wang, Lei and Yang, Min and Xu, Ruifeng},
  booktitle={Findings of the Association for Computational Linguistics: ACL 2024},
  pages={14268--14290},
  year={2024}
}

@article{he2021chemu,
  title={Chemu 2020: Natural language processing methods are effective for information extraction from chemical patents},
  author={He, Jiayuan and Nguyen, Dat Quoc and Akhondi, Saber A and Druckenbrodt, Christian and Thorne, Camilo and Hoessel, Ralph and Afzal, Zubair and Zhai, Zenan and Fang, Biaoyan and Yoshikawa, Hiyori and others},
  journal={Frontiers in Research Metrics and Analytics},
  volume={6},
  pages={654438},
  year={2021},
  publisher={Frontiers Media SA}
}

@inproceedings{zheng2024llamafactory,
  title={LlamaFactory: Unified Efficient Fine-Tuning of 100+ Language Models},
  author={Yaowei Zheng and Richong Zhang and Junhao Zhang and Yanhan Ye and Zheyan Luo and Zhangchi Feng and Yongqiang Ma},
  booktitle={Proceedings of the 62nd Annual Meeting of the Association for Computational Linguistics (Volume 3: System Demonstrations)},
  address={Bangkok, Thailand},
  publisher={Association for Computational Linguistics},
  year={2024},
  url={http://arxiv.org/abs/2403.13372}
}

@misc{unsloth,
  author = {Daniel Han, Michael Han and Unsloth team},
  title = {Unsloth},
  url = {https://github.com/unslothai/unsloth},
  year = {2023}
}

@inproceedings{harper-etal-2021-semeval,
    title = "{S}em{E}val-2021 Task 8: {M}eas{E}val {--} Extracting Counts and Measurements and their Related Contexts",
    author = "Harper, Corey  and
      Cox, Jessica  and
      Kohler, Curt  and
      Scerri, Antony  and
      Daniel Jr., Ron  and
      Groth, Paul",
    editor = "Palmer, Alexis  and
      Schneider, Nathan  and
      Schluter, Natalie  and
      Emerson, Guy  and
      Herbelot, Aurelie  and
      Zhu, Xiaodan",
    booktitle = "Proceedings of the 15th International Workshop on Semantic Evaluation (SemEval-2021)",
    month = aug,
    year = "2021",
    address = "Online",
    publisher = "Association for Computational Linguistics",
    url = "https://aclanthology.org/2021.semeval-1.38/",
    doi = "10.18653/v1/2021.semeval-1.38",
    pages = "306--316"
}

@inproceedings{davletov-etal-2021-liori-semeval,
    title = "{LIORI} at {S}em{E}val-2021 Task 8: Ask Transformer for measurements",
    author = "Davletov, Adis  and
      Gordeev, Denis  and
      Arefyev, Nikolay  and
      Davletov, Emil",
    editor = "Palmer, Alexis  and
      Schneider, Nathan  and
      Schluter, Natalie  and
      Emerson, Guy  and
      Herbelot, Aurelie  and
      Zhu, Xiaodan",
    booktitle = "Proceedings of the 15th International Workshop on Semantic Evaluation (SemEval-2021)",
    month = aug,
    year = "2021",
    address = "Online",
    publisher = "Association for Computational Linguistics",
    url = "https://aclanthology.org/2021.semeval-1.178/",
    doi = "10.18653/v1/2021.semeval-1.178",
    pages = "1249--1254"
}

@inproceedings{cao-etal-2021-conner,
    title = "{CONNER}: A Cascade Count and Measurement Extraction Tool for Scientific Discourse",
    author = "Cao, Jiarun  and
      Xiang, Yuejia  and
      Zhang, Yunyan  and
      Qi, Zhiyuan  and
      Chen, Xi  and
      Zheng, Yefeng",
    editor = "Palmer, Alexis  and
      Schneider, Nathan  and
      Schluter, Natalie  and
      Emerson, Guy  and
      Herbelot, Aurelie  and
      Zhu, Xiaodan",
    booktitle = "Proceedings of the 15th International Workshop on Semantic Evaluation (SemEval-2021)",
    month = aug,
    year = "2021",
    address = "Online",
    publisher = "Association for Computational Linguistics",
    url = "https://aclanthology.org/2021.semeval-1.176/",
    doi = "10.18653/v1/2021.semeval-1.176",
    pages = "1239--1244"
}

@inproceedings{gangwar-etal-2021-counts,
    title = "Counts@{IITK} at {S}em{E}val-2021 Task 8: {S}ci{BERT} Based Entity And Semantic Relation Extraction For Scientific Data",
    author = "Gangwar, Akash  and
      Jain, Sabhay  and
      Sourav, Shubham  and
      Modi, Ashutosh",
    editor = "Palmer, Alexis  and
      Schneider, Nathan  and
      Schluter, Natalie  and
      Emerson, Guy  and
      Herbelot, Aurelie  and
      Zhu, Xiaodan",
    booktitle = "Proceedings of the 15th International Workshop on Semantic Evaluation (SemEval-2021)",
    month = aug,
    year = "2021",
    address = "Online",
    publisher = "Association for Computational Linguistics",
    url = "https://aclanthology.org/2021.semeval-1.175/",
    doi = "10.18653/v1/2021.semeval-1.175",
    pages = "1232--1238"
}

@inproceedings{liu-etal-2024-reactxt,
    title = "{R}eact{XT}: Understanding Molecular ``Reaction-ship'' via Reaction-Contextualized Molecule-Text Pretraining",
    author = "Liu, Zhiyuan  and
      Shi, Yaorui  and
      Zhang, An  and
      Li, Sihang  and
      Zhang, Enzhi  and
      Wang, Xiang  and
      Kawaguchi, Kenji  and
      Chua, Tat-Seng",
    editor = "Ku, Lun-Wei  and
      Martins, Andre  and
      Srikumar, Vivek",
    booktitle = "Findings of the Association for Computational Linguistics: ACL 2024",
    month = aug,
    year = "2024",
    address = "Bangkok, Thailand",
    publisher = "Association for Computational Linguistics",
    url = "https://aclanthology.org/2024.findings-acl.318/",
    doi = "10.18653/v1/2024.findings-acl.318",
    pages = "5353--5377"
}

@inproceedings{almasian-etal-2023-cqe,
    title = "{CQE}: A Comprehensive Quantity Extractor",
    author = {Almasian, Satya  and
      Kazakova, Vivian  and
      G{\"o}ldner, Philipp  and
      Gertz, Michael},
    editor = "Bouamor, Houda  and
      Pino, Juan  and
      Bali, Kalika",
    booktitle = "Proceedings of the 2023 Conference on Empirical Methods in Natural Language Processing",
    month = dec,
    year = "2023",
    address = "Singapore",
    publisher = "Association for Computational Linguistics",
    url = "https://aclanthology.org/2023.emnlp-main.793/",
    doi = "10.18653/v1/2023.emnlp-main.793",
    pages = "12845--12859"
}
\newpage
\appendix
\definecolor{customdarkgray}{RGB}{60, 60, 60}

\newtcolorbox{promptbox}[1]{
    enhanced,              
    colback=white,         
    colframe=customdarkgray, 
    colbacktitle=customdarkgray, 
    coltitle=white,        
    fonttitle=\bfseries\large,
    fontupper=\small\rmfamily, 
    title={#1},            
    arc=1mm,               
    boxrule=1.5pt,         
    titlerule=0mm,         
    toptitle=2mm,
    bottomtitle=2mm,
    boxsep=2mm,
    sharp corners=south,   
    before skip=0pt,       
    after skip=0pt,        
    left=2pt, right=2pt,   
}

\section{Prompt template}
\label{sec:prompt}

\noindent
\begin{minipage}[t]{\textwidth}

\begin{minipage}[t]{0.48\textwidth}
    \vspace{0pt} 
    
    \begin{promptbox}{Prompt for $\mathcal{P}_{\text{trace}}$}
        \textbf{Instruction:} \\
        You are an expert in extracting structured annotations from text.
        I have an text input and you need to extract all the quantities within it. I need you to strictly follow the format with six specific sections: ARABIC-QUANTITY, NUMERIC-QUANTITY, TIME-QUANTITY, CHANGE-QUANTITY, CHANGE-QUANTITY, FORMULA-QUANTITY, CONCLUSION. 
        
        To explain further: In ARABIC-QUANTITY, outline a step-by-step thought process you use to extract quantity in arabic form. In NUMERIC-QUANTITY, outline a step by step thought process \dots In CONCLUSION, give the final answer in a tsv format explained below.
        
        I will provide you with the quantities extracted using the quantulum library for your reference, the information provided by Quantulum is standardized. You need to find the original text in the passage and fill in the tsv form. Also, the quantulum information maybe incorrect, You can't follow it completely.
        
        \vspace{1em}
        Here's how the format should look: <ARABIC-QUANTITY> [Provide a chain-of-thought explanation of how you extract all quantities in the arabic forms] </ARABIC-QUANTITY> <NUMERIC-QUANTITY> \dots <CONCLUSION>[State the final answer in a tsv format explained below format\dots] </CONCLUSION>
        
        \vspace{1em}
        \textbf{Task Definition: Extract Quantities}
        
        1. Annotation of Quantities: \dots
        
        2. Example Process: \dots
        
        Output Format (TSV Fields): \dots
        
        Final Output Example: \dots
        
        \textbf{The reference answer from quantulum:} \dots
    \end{promptbox}
\end{minipage}%
\hfill
\begin{minipage}[t]{0.48\textwidth}
    \vspace{0pt} 
    
    \begin{promptbox}{Prompt for $\mathcal{P}_{\text{aug}}$}
        \textbf{Instruction:} \\
        You are an expert in extracting structured annotations from text.
        I have an text input and you need to extract all the quantities within it. I need you to strictly follow the format with six specific sections: ARABIC-QUANTITY, NUMERIC-QUANTITY, TIME-QUANTITY, CHANGE-QUANTITY, CHANGE-QUANTITY, FORMULA-QUANTITY, CONCLUSION. 
        
        To explain further: In ARABIC-QUANTITY, outline a step-by-step thought process you use to extract quantity in arabic form. In NUMERIC-QUANTITY, outline a step by step thought process \dots In CONCLUSION, give the final answer in a tsv format explained below. It is crucial that you adhere to this structure exactly as outlined and that the final answer in the CONCLUSION matches the standard correct answer precisely.
        
        \vspace{1em}
        Here's how the format should look: <ARABIC-QUANTITY> [Provide a chain-of-thought explanation of how you extract all quantities in the arabic forms] </ARABIC-QUANTITY> <NUMERIC-QUANTITY> \dots <CONCLUSION>[State the final answer in a tsv format explained below format\dots It is crucial that you adhere to this structure exactly as outlined and that the final answer] </CONCLUSION>
        
        \vspace{1em}
        \textbf{Task Definition: Extract Quantities}
        
        1. Annotation of Quantities: \dots
        
        2. Example Process: \dots
        
        Output Format (TSV Fields): \dots
        
        Final Output Example: \dots
        
        \textbf{The gold answers:} \dots
    \end{promptbox}
\end{minipage}
\end{minipage}

\twocolumn

\section{Case Study: High-Entropy Token Suppression by GRPO}
\label{app:case_study}

To better understand the effect of GRPO, we examine a representative sample:

\paragraph{Input:}
\begin{quote}
Samples were then annealed in air in a pre-heated furnace at temperatures up to 798\,°C for times chosen to ensure complete iron diffusion through the sample.
\end{quote}

\paragraph{Gold Standard Annotation:}
\begin{itemize}
    \item \textbf{Quantity:} \texttt{up to 798\,°C}  
      \item    \textbf{Unit:} \texttt{°C}  
       \item   \textbf{Modifiers:} \texttt{["IsRange"]}
    \item \textbf{MeasuredProperty:} \texttt{annealed in air in a pre-heated furnace at temperatures}
    \item \textbf{MeasuredEntity:} \texttt{Samples}
\end{itemize}

\paragraph{SFT Output:}
\begin{quote}
We can find the quantity with surface form [up to 798\,°C], it has unit [°C]. The modifier for the quantity are [IsRange]. This quantity is used to describe the entity [\textcolor{red}{f}urnace]. The entity has the following property [temperatures].
\end{quote}

Here, the red token indicates a high-entropy token (entropy = 1.39), and the top-5 candidate tokens with probabilities are listed in the first column of Table~\ref{tab:token_comparison}.

\paragraph{GRPO Output:}
\begin{quote}
We can find the quantity with surface form [up to 798\,°C], it has unit [°C]. The modifier for the quantity are [IsRange]. This quantity is used to describe the entity [Samples]. The entity has the following property [temperatures].
\end{quote}

There is no high-entropy token in the model's output after the GRPO training. The top-5 candidate tokens with probabilities at the [Samples] position are shown in the second column of Table~\ref{tab:token_comparison}.

\paragraph{Analysis:}
GRPO successfully suppresses the high-entropy token observed in SFT, assigning the correct token \texttt{Samples} with high confidence and eliminating uncertainty, demonstrating improved reasoning stability and more deterministic output.
\begin{table}[ht!]
\centering
\footnotesize
\begin{tabular}{c|lc|lc}
\toprule
& \multicolumn{2}{c|}{\textbf{w/o GRPO}} & \multicolumn{2}{c}{\textbf{GRPO}} \\
\textbf{Rank} & \textbf{Candidate} & \textbf{Prob} & \textbf{Candidate} & \textbf{Prob} \\
\midrule
1 & \textcolor{red}{f} & 0.547 & \textbf{Samples} & 0.847 \\
2 & Samples & 0.376 & ` Samples & 0.115 \\
3 & ` Samples & 0.051 & f & 0.037 \\
4 & samples & 0.015 & samples & 0.0013 \\
5 & pre & 0.011 & \_samples & 0.0001 \\
\bottomrule
\end{tabular}
\caption{Comparison of Top-5 candidate tokens at the target position between SFT and GRPO outputs.}
\label{tab:token_comparison}
\end{table}

\section{MeasEval-Ext and its Annotation Details}
\label{sec:annotation_details}

The annotations are drawn from recent research papers that postdate the original MeasEval corpus. A distinct advantage of this data source is its adversarial selection strategy: unlike the randomized distribution in the original dataset, we deliberately curated 135 text segments(the same as the MeasEval evaluation dataset) containing \textbf{novel units and complex quantity expressions absent from the training distribution}. This design ensures that the dataset strictly tests the model's ability to identify and ground quantities based on semantic context rather than memorized vocabulary.

To ensure high data quality, we enlisted researchers from our laboratory as annotators. The annotation process strictly followed the official \textit{MeasEval Annotation Guidelines}. All samples were \textbf{independently labeled by two annotators} to capture the dense quantity-centric information. Following the initial annotation, results were reviewed and reconciled during an \textbf{adjudication meeting} to resolve disagreements and reach a final consensus.

The consistency of the dataset is validated by the Inter-Annotator Agreement (IAA). As shown in Table~\ref{tab:k_alpha_scores}, the Krippendorff's Alpha scores (e.g., 0.921 for Quantity) indicate strong agreement, comparable to the original MeasEval benchmarks.

\begin{table}[h] 
    \centering

    \begin{tabular}{lc}
        \toprule
        \textbf{Annotation Class} & \textbf{Krippendorff's $\alpha$} \\
        \midrule
        Quantity & 0.921 \\
        MeasuredEntity & 0.639 \\
        MeasuredProperty & 0.584 \\
        Qualifier & 0.416 \\
        \bottomrule
    \end{tabular}
        \caption{Inter-Annotator agreement (Krippendorff's Alpha) for MeasEval-Ext}
    \label{tab:k_alpha_scores}
\end{table}

\section{Quantity Phase Reward}
\label{sec:qty_reward}

 The total reward $R(y)$ is a weighted sum of four components for mitigating distinct Quantity hallucination types, which includes the Format Reward ($r_{\text{fmt}}$), the Out-of-Scope Penalty ($r_{\text{scope}}$), the Fabrication Penalty ($r_{\text{fab}}$) and the Misclassification Penalty ($r_{\text{mis}}$):
\begin{equation}
    R(y) = r_{\text{fmt}}(y) + r_{\text{scope}}(y) + r_{\text{fab}}(y) + r_{\text{mis}}(y)
\end{equation}

\begin{enumerate}
    \item[\textbf{$r_{\text{fmt}}$}]
        To enforce output schema compliance, we validate the sequential semantic tags $\mathcal{S}_{\text{tags}} = \{\texttt{<ARABIC>}, \dots, \texttt{<CONCLUSION>}\}$ via regex pattern $\mathcal{P}_{\text{struct}}$. The binary reward is:
        \begin{equation}
            r_{\text{fmt}}(y) = \mathbbm{I}\Big( y \equiv \mathcal{P}_{\text{struct}} \Big)
        \end{equation}
\item[\textbf{$r_{\text{scope}}$}] \raggedright
    Constrains out-of-scope entities(e.g., ``Fig.~1'') via local patterns $\mathcal{C}(e)$ and global precision $P_{\text{ans}}$:
    \vspace{-0.5em}
    \begin{equation}
        r_{\text{scope}} = -\lambda_{\text{loc}} \sum_{e} \mathcal{C}(e) + \beta_{\text{scope}} P_{\text{ans}}
    \end{equation}
    \item[\textbf{$r_{\text{fab}}$}]
        Prohibiting invalid quantity fabrication via parsers combining CQE\citep{almasian-etal-2023-cqe} and Quantulum\footnote{\url{https://github.com/nielstron/quantulum3}} $\mathcal{T}_{\text{parse}}$, the penalty includes grounding constraints:
        \begin{align}
            r_{\text{fab}}(y) &=  -\lambda_{\text{fab}}\sum_{e \in \mathcal{E}_y} \mathbbm{I}\Big( \mathcal{T}_{\text{parse}}(e)=\emptyset \Big)
        \end{align}

    \item[\textbf{$r_{\text{mis}}$}]
        Mitigating span boundary errors via token-level precision $P_{\text{tok}}$, the reward is:
        \begin{equation}
            r_{\text{mis}}(y) = \bar{F1}_{\text{tok}} - \lambda_{\text{mis}} \cdot (1 - P_{\text{tok}})
        \end{equation}
\end{enumerate}

\section{Relation Phase Reward}
\label{sec:rel_reward}

While the sentence-based extraction excels at local entity identification (e.g., Units, Modifiers), it suffers from failures in capturing long-range dependency chains (e.g., \textit{MeasuredEntity}, \textit{MeasuredProperty}, \textit{Qualifier}), inference bias, and under-extraction of sparse components. 
To address these issues, enforce logical completeness, suppress Quantity hallucinations, and incentivize sparse component retrieval, we design a composite reward function $R(y)$ optimized via GRPO. 
The total reward is a weighted sum of three dedicated components~(Format \& Grounding Reward ($r_{\text{fmt}}$)), Relation Completeness \& Exploration Reward ($r_{\text{comp}}$), and Misclassification Penalty ($r_{\text{mis}}$)), that target distinct quantitative extraction flaws:
\begin{equation}
    R(y) = r_{\text{fmt}}(y) + r_{\text{comp}}(y) + r_{\text{mis}}(y)
\end{equation}

\noindent The reward components are elaborated as follows with explicit optimization objectives and mathematical formulations:
\begin{enumerate}
\item[\textbf{$r_{\text{fmt}}$}]
    To enforce structural consistency, we validate the existence of analysis sections $\mathcal{S}_y$ and adherence to the SFT schema $\mathcal{F}_{\text{SFT}}$. The binary reward is:
    \begin{equation}
        r_{\text{fmt}}(y) = \mathbbm{I}\Big( \mathcal{S}_y \neq \emptyset \land y \models \mathcal{F}_{\text{SFT}} \Big)
    \end{equation}

\item[\textbf{$r_{\text{comp}}$}] \raggedright
    Drives \textbf{comprehensive exploration} by aligning predicted groups $p$ with gold groups $g$. To prevent partial extraction, we incentivize full recovery via stepwise matching, closure bonuses, and weighted component bonuses:
    \vspace{-0.5em}
    \begin{equation}
    \begin{split}
        r_{\text{comp}}(y) &= \sum_{p \sim g} \Big( \underbrace{\lambda_{\text{step}} |p \cap g|}_{\text{Stepwise}} + \underbrace{\beta_{\text{full}} \mathbbm{I}(g \subseteq p)}_{\text{Closure}} \Big) \\
        &+ \underbrace{\lambda_{\text{exp}} \sum\nolimits_{c} w_c F1_c^{\text{ans}}}_{\text{Weighted Exploration}}
    \end{split}
    \end{equation}
    \noindent where weights $\mathbf{w}$ prioritize harder-to-predict dependencies to ensure no critical node is missed.

\item[\textbf{$r_{\text{mis}}$}] \raggedright
    Suppresses over-broad spans by penalizing token-level precision loss $(1 - P_{\text{tok}})$:
    \vspace{-0.5em}
    \begin{equation}
        r_{\text{mis}}(y) = F1_{\text{tok}} - (1 - P_{\text{tok}})
    \end{equation}    

\end{enumerate}

\begin{table*}[htbp]
    \centering
    \small 

    \renewcommand{\arraystretch}{1.5} 
    
    \begin{tabularx}{\textwidth}{@{}l X X X X X@{}}
        \toprule
        \textbf{Type} & \textbf{Fabrication} & \textbf{Out-of-Scope} & \textbf{Misclassification} & \textbf{Inference Bias} & \textbf{Role Definition} \\ 
        \midrule
        
        \textbf{Quantity} & 
        Generates fictitious values absent from the source text, or yields extracted strings that contain no valid numerical content. & 
        Extracts invalid numerical tokens, including figure citations (e.g., ``Fig. 4'') or scientific nomenclature containing digits (e.g., ``4S RNA''). & 
        Generates excessively long spans that erroneously incorporate surrounding components, such as the \textit{MeasuredEntity}. & 
        \multicolumn{1}{c}{--} & 
        \multicolumn{1}{c}{--} \\ 
        
        \textbf{Relation} & 
        \multicolumn{1}{c}{--} & 
        \multicolumn{1}{c}{--} & 
        Generates excessively long spans that erroneously incorporate surrounding context or unrelated text segments. & 
        Propagates errors from preceding components (e.g., incorrect \textit{MeasuredEntity} will result in  cascading hallucinations in properties and qualifiers). & 
        Fails to distinguish semantic roles, frequently inverting the \textit{MeasuredEntity} and \textit{MeasuredProperty}. \\ 
        
        \bottomrule
    \end{tabularx}
        \caption{Taxonomy of Hallucinations in Information Extraction}
    \label{tab:hallucination_taxonomy}
\end{table*}

\section{Efficiency Analysis of Sentence-Based Reasoning}
\label{sec:efficiency}
To analyze the efficiency of the proposed sentence-based reasoning strategy, we compare it with the rule-based method in terms of token consumption.

On the MeasEval benchmark, the sentence-based approach requires an average of 871 tokens per response, which is a 27\% reduction compared to the 1,193 tokens used by the rule-based method. This indicates that our method improves inference efficiency rather than introducing additional overhead.

This efficiency gain can be attributed to the use of local context anchoring. While rule-based approaches often rely on constructing global reasoning chains across the entire input, our method focuses on extracting relevant elements from localized sentence-level contexts, where most measurement relations are concentrated. This reduces redundant reasoning paths and avoids unnecessary global search, leading to more efficient inference without compromising performance.

\section{Generalization to Unseen Domains}
\label{sec:generalization}
To evaluate whether our method relies on dataset-specific priors or learns generalizable hallucination mitigation capabilities for extraction, we further assess \textsc{MeasHalu} on an unseen-domain benchmark, ChEMU-NER~\cite{he2021chemu}, without additional fine-tuning.

The ChEMU dataset consists of chemical patent texts annotated with entity roles such as reaction products, starting materials, and experimental conditions (e.g., time, temperature, and yield). These elements exhibit partial semantic alignment with the MeasEval schema, enabling evaluation via direct schema mapping.

We compare two prompt strategies: (1) \textit{Task-Specific Optimized}, where prompts are tailored to the chemical domain, and (2) \textit{MeasEval Style}, where models are evaluated using our generalized extraction framework without domain-specific adaptation.

\begin{table}[t]
\centering
\setlength{\tabcolsep}{3pt}
\footnotesize
\resizebox{\columnwidth}{!}{
\begin{tabular}{l l c c c}
\toprule
\textbf{Schema} & \textbf{Model} & \textbf{P} & \textbf{R} & \textbf{F1} \\
\midrule

\multirow{2}{*}{Task-Specific}
& DeepSeek-R1 & \valueWithStd{0.7200}{0.0014} & \valueWithStd{0.7835}{0.0120} & \valueWithStd{0.7505}{0.0049} \\
& DeepSeek-V3 & \valueWithStd{0.6850}{0.0165} & \valueWithStd{0.7433}{0.0418} & \valueWithStd{0.7123}{0.0120} \\

\midrule

\multirow{4}{*}{MeasEval}
& MeasHalu & \textbf{\valueWithStd{0.8465}{0.0021}} & \textbf{\valueWithStd{0.6845}{0.0035}} & \textbf{\valueWithStd{0.7570}{0.0028}} \\
& GPT-5 & \valueWithStd{0.5317}{0.0060} & \valueWithStd{0.6583}{0.0915} & \valueWithStd{0.5857}{0.0344} \\
& DeepSeek-R1 & \valueWithStd{0.5533}{0.0085} & \valueWithStd{0.4993}{0.0246} & \valueWithStd{0.5243}{0.0152} \\
& DeepSeek-V3 & \valueWithStd{0.5527}{0.0068} & \valueWithStd{0.3807}{0.0101} & \valueWithStd{0.4507}{0.0061} \\

\bottomrule
\end{tabular}
}
\caption{Performance comparison under different prompt strategies.}
\label{tab:prompt_strategy_comparison}
\end{table}

As shown in Table~\ref{tab:prompt_strategy_comparison}, MeasHalu achieves the best F1 score (0.7570) under the MeasEval-style setting, outperforming strong baselines such as DeepSeek-R1 and DeepSeek-V3. Notably, MeasHalu attains significantly higher precision (0.8465), indicating strong robustness in suppressing hallucinated predictions in unseen domains.

Although recall is lower due to schema mismatch and incomplete mapping, the overall performance demonstrates that \textsc{MeasHalu} learns transferable hallucination mitigation capabilities for extraction rather than relying on dataset-specific priors.

\section{Analysis of Sparse Component Extraction under Relaxed Matching}
\label{sec:qualifier_analysis}

Extracting sparse elements such as qualifiers remains a challenging problem in scientific measurement extraction. This difficulty largely stems from the ambiguous semantic dependencies in scientific texts, where such elements often lack clear syntactic boundaries and exhibit high variability in expression.

Under strict span-based evaluation (e.g., Strict Overlap F1), even minor boundary deviations can lead to substantial performance penalties, despite the model correctly identifying the core semantic region. For example, partial span mismatches (e.g., ``with respect to earth'' vs. ``respect to earth'') are treated as complete errors.

To better reflect semantic correctness, we further evaluate model performance using a relaxed matching criterion following the MeasEval protocol, which allows partial span overlap. Under this metric, MeasHalu-7B achieves an F1 score of 0.315, indicating strong localization capability despite boundary ambiguity.

\begin{table}[h]
\centering
\small
\begin{tabular}{lccc}
\toprule
Model & Precision & Recall & F1 \\
\midrule
MeasHalu-7B & 0.436 & 0.246 & \textbf{0.315} \\
\multicolumn{1}{r}{w/o GRPO} & 0.325 & 0.259 & 0.288 \\
GPT-5 & 0.220 & 0.348 & 0.269 \\
Gemini-2.5-Pro & 0.225 & 0.360 & 0.277 \\
DeepSeek-R1 & 0.269 & 0.202 & 0.231 \\
\bottomrule
\end{tabular}
\caption{Performance under relaxed matching for sparse components.}
\end{table}

These results suggest that, although strict span-based metrics indicate low performance on sparse elements, the model is capable of accurately localizing the relevant semantic regions. Furthermore, the relatively high precision demonstrates effective suppression of hallucinations even under challenging extraction settings.

\begin{table*}[t]
\centering
\setlength{\tabcolsep}{1pt} 
\scriptsize
    \begin{tabular}{l c c ccc c cc c cc c cc} 
    \toprule
    \multirow{3}{*}{\textbf{Model}} & \multirow{3}{*}{\textbf{Overall}} & & \multicolumn{3}{c}{\textbf{Quantities}} & & \multicolumn{2}{c}{\textbf{Entities}} & & \multicolumn{2}{c}{\textbf{Properties}} & & \multicolumn{2}{c}{\textbf{Qualifiers}} \\
    \cmidrule{4-6}\cmidrule{8-9}\cmidrule{11-12}\cmidrule{14-15}
     & & & \textbf{Quantity} & \textbf{Unit} & \textbf{Modifier} & & \textbf{ME} & \textbf{HasQuantity} & & \textbf{MP} & \textbf{HasProperty} & & \textbf{Qualifier} & \textbf{Qualifies} \\
    \midrule    
    \multicolumn{15}{c}{\textit{Top Ranked Systems from the MeasEval Competition}} \\ 
    \midrule
    Baseline & 0.239 & & 0.827 & 0.561 & 0.000 & & 0.053 & 0.075 & & 0.064 & 0.007 & & 0.005 & 0.000 \\
    Counts & 0.432 & & \cellcolor{lightred}{0.861} & 0.804 & \cellcolor{lightyellow}{0.614} & & 0.406 & 0.311 & & 0.245 & 0.183 & & 0.077 & 0.064 \\
    CONNER & 0.473 & & \cellcolor{lightyellow}{0.855} & 0.719 & 0.523 & & 0.398 & 0.424 & & \cellcolor{lightteal}{0.437} & 0.257 & & 0.000 & 0.000 \\
    LIORI & \cellcolor{lightred}{0.519} & & \cellcolor{lightred}{0.861} & 0.722 & \cellcolor{lightred}{0.642} & & \cellcolor{lightyellow}{0.437} & \cellcolor{lightred}{0.482} & & \cellcolor{lightred}{0.467} & \cellcolor{lightred}{0.318} & & \cellcolor{lightyellow}{0.163} & \cellcolor{lightyellow}{0.092} \\
    \midrule
    \multicolumn{15}{c}{\textit{Rule-based Prompting}} \\ 
    \midrule
    Qwen2.5-7b-inst &\valueWithStd{0.171}{0.028}&&\valueWithStd{0.491}{0.052}&\valueWithStd{0.478}{0.075}&\valueWithStd{0.106}{0.011}&&\valueWithStd{0.088}{0.021}&\valueWithStd{0.045}{0.015}&&\valueWithStd{0.057}{0.008}&\valueWithStd{0.000}{0.000}&&\valueWithStd{0.040}{0.012}&\valueWithStd{0.017}{0.006}\\    
    Qwen2.5-72b-inst &\valueWithStd{0.286}{0.028}&&\valueWithStd{0.644}{0.035}&\valueWithStd{0.826}{0.026}&\valueWithStd{0.236}{0.115}&&\valueWithStd{0.196}{0.038}&\valueWithStd{0.147}{0.028}&&\valueWithStd{0.164}{0.042}&\valueWithStd{0.001}{0.002}&&\valueWithStd{0.076}{0.019}&\valueWithStd{0.021}{0.011}\\
    DeepSeek-R1 &\valueWithStd{0.253}{0.008}&&\valueWithStd{0.569}{0.002}&\valueWithStd{0.586}{0.002}&\valueWithStd{0.240}{0.018}&&\valueWithStd{0.216}{0.017}&\valueWithStd{0.163}{0.014}&&\valueWithStd{0.163}{0.003}&\valueWithStd{0.024}{0.010}&&\valueWithStd{0.085}{0.015}&\valueWithStd{0.029}{0.002}\\
    DeepSeek-V3 &\valueWithStd{0.271}{0.008}&&\valueWithStd{0.657}{0.018}&\valueWithStd{0.768}{0.010}&\valueWithStd{0.214}{0.003}&&\valueWithStd{0.239}{0.009}&\valueWithStd{0.135}{0.014}&&\valueWithStd{0.113}{0.011}&\valueWithStd{0.001}{0.002}&&\valueWithStd{0.085}{0.004}&\valueWithStd{0.014}{0.004}\\
    Gemini-2.5-Pro &\valueWithStd{0.359}{0.008}&&\valueWithStd{0.712}{0.007}&\valueWithStd{0.784}{0.035}&\valueWithStd{0.464}{0.009}&&\valueWithStd{0.306}{0.009}&\valueWithStd{0.266}{0.007}&&\valueWithStd{0.287}{0.021}&\valueWithStd{0.090}{0.009}&&\valueWithStd{0.146}{0.018}&\cellcolor{lightteal}{\valueWithStd{0.076}{0.014}}\\
    GPT-5 &\valueWithStd{0.371}{0.004}&&\valueWithStd{0.804}{0.013}&\valueWithStd{0.742}{0.018}&\valueWithStd{0.395}{0.020}&&\valueWithStd{0.361}{0.003}&\valueWithStd{0.270}{0.005}&&\valueWithStd{0.355}{0.014}&\valueWithStd{0.020}{0.004}&&\valueWithStd{0.152}{0.019}&\valueWithStd{0.052}{0.006}\\
    \midrule
    \multicolumn{15}{c}{\textit{Sentence-based Prompting}} \\
    \midrule
    Qwen2.5-7b-inst &\valueWithStd{0.073}{0.006}&&\valueWithStd{0.151}{0.008}&\valueWithStd{0.160}{0.007}&\valueWithStd{0.027}{0.001}&&\valueWithStd{0.066}{0.009}&\valueWithStd{0.059}{0.010}&&\valueWithStd{0.044}{0.010}&\valueWithStd{0.031}{0.010}&&\valueWithStd{0.003}{0.004}&\valueWithStd{0.005}{0.006}\\    
    Qwen2.5-72b-inst &\valueWithStd{0.212}{0.002}&&\valueWithStd{0.403}{0.006}&\valueWithStd{0.516}{0.005}&\valueWithStd{0.232}{0.010}&&\valueWithStd{0.204}{0.005}&\valueWithStd{0.137}{0.007}&&\valueWithStd{0.113}{0.008}&\valueWithStd{0.087}{0.005}&&\valueWithStd{0.038}{0.011}&\valueWithStd{0.012}{0.007}\\
    DeepSeek-R1 &\valueWithStd{0.304}{0.004}&&\valueWithStd{0.589}{0.016}&\valueWithStd{0.711}{0.021}&\valueWithStd{0.356}{0.030}&&\valueWithStd{0.260}{0.011}&\valueWithStd{0.198}{0.012}&&\valueWithStd{0.182}{0.006}&\valueWithStd{0.118}{0.008}&&\valueWithStd{0.113}{0.030}&\valueWithStd{0.057}{0.019}\\  
    DeepSeek-V3 &\valueWithStd{0.320}{0.006}&&\valueWithStd{0.567}{0.013}&\valueWithStd{0.726}{0.005}&\valueWithStd{0.355}{0.016}&&\valueWithStd{0.303}{0.006}&\valueWithStd{0.225}{0.018}&&\valueWithStd{0.226}{0.022}&\valueWithStd{0.149}{0.002}&&\valueWithStd{0.019}{0.009}&\valueWithStd{0.000}{0.000}\\    
    Gemini-2.5-Pro &\valueWithStd{0.440}{0.011}&&\valueWithStd{0.782}{0.003}&\cellcolor{lightred}{\valueWithStd{0.882}{0.003}}&\valueWithStd{0.486}{0.011}&&\cellcolor{lightteal}{\valueWithStd{0.436}{0.017}}&\valueWithStd{0.376}{0.020}&&\valueWithStd{0.386}{0.028}&\cellcolor{lightteal}{\valueWithStd{0.280}{0.025}}&&\valueWithStd{0.143}{0.019}&\valueWithStd{0.056}{0.010}\\
    GPT-5 &\valueWithStd{0.406}{0.008}&&\valueWithStd{0.724}{0.007}&\valueWithStd{0.817}{0.017}&\valueWithStd{0.500}{0.027}&&\valueWithStd{0.397}{0.002}&\valueWithStd{0.351}{0.006}&&\valueWithStd{0.355}{0.009}&\valueWithStd{0.226}{0.002}&&\valueWithStd{0.138}{0.026}&\valueWithStd{0.042}{0.031}\\    
    \midrule
    MeasHalu-0.5B&\valueWithStd{0.347}{0.003}&&\valueWithStd{0.659}{0.009}&\valueWithStd{0.760}{0.015}&\valueWithStd{0.268}{0.007}&&\valueWithStd{0.284}{0.002}&\valueWithStd{0.279}{0.006}&&\valueWithStd{0.287}{0.006}&\valueWithStd{0.184}{0.003}&&\valueWithStd{0.059}{0.017}&\valueWithStd{0.037}{0.010}\\
     \multicolumn{1}{r}{w/o GRPO}  &\valueWithStd{0.312}{0.008}&&\valueWithStd{0.649}{0.001}&\valueWithStd{0.717}{0.019}&\valueWithStd{0.239}{0.006}&&\valueWithStd{0.263}{0.004}&\valueWithStd{0.241}{0.017}&&\valueWithStd{0.238}{0.017}&\valueWithStd{0.140}{0.004}&&\valueWithStd{0.069}{0.009}&\valueWithStd{0.022}{0.007}\\
    MeasHalu-3B &\valueWithStd{0.460}{0.005}&&\valueWithStd{0.810}{0.004}&\cellcolor{lightteal}{\valueWithStd{0.875}{0.005}}&\valueWithStd{0.440}{0.002}&&\valueWithStd{0.396}{0.007}&\valueWithStd{0.418}{0.007}&&\valueWithStd{0.407}{0.006}&\valueWithStd{0.277}{0.007}&&\valueWithStd{0.110}{0.005}&\valueWithStd{0.049}{0.010}\\
     \multicolumn{1}{r}{w/o GRPO}  &\valueWithStd{0.433}{0.010}&&\valueWithStd{0.782}{0.014}&\valueWithStd{0.850}{0.003}&\valueWithStd{0.449}{0.009}&&\valueWithStd{0.370}{0.006}&\valueWithStd{0.377}{0.017}&&\valueWithStd{0.365}{0.019}&\valueWithStd{0.245}{0.002}&&\valueWithStd{0.084}{0.006}&\valueWithStd{0.043}{0.015}\\
    MeasHalu-7B &\cellcolor{lightyellow}{\valueWithStd{0.512}{0.004}}&&\cellcolor{lightteal}{\valueWithStd{0.848}{0.001}}&\valueWithStd{0.860}{0.008}&\valueWithStd{0.607}{0.006}&&\cellcolor{lightred}{\valueWithStd{0.455}{0.008}}&\cellcolor{lightyellow}{\valueWithStd{0.472}{0.009}}&&\cellcolor{lightyellow}{\valueWithStd{0.442}{0.012}}&\cellcolor{lightyellow}{\valueWithStd{0.310}{0.005}}&&\cellcolor{lightred}{\valueWithStd{0.170}{0.005}}&\cellcolor{lightred}{\valueWithStd{0.100}{0.009}}\\
     \multicolumn{1}{r}{w/o GRPO} &\cellcolor{lightteal}{\valueWithStd{0.479}{0.005}}&&\valueWithStd{0.846}{0.002}&\cellcolor{lightyellow}{\valueWithStd{0.863}{0.004}}&\cellcolor{lightteal}{\valueWithStd{0.610}{0.015}}&&\valueWithStd{0.429}{0.004}&\cellcolor{lightteal}{\valueWithStd{0.433}{0.016}}&&\valueWithStd{0.397}{0.016}&\valueWithStd{0.272}{0.012}&&\cellcolor{lightteal}{\valueWithStd{0.155}{0.012}}&\valueWithStd{0.063}{0.010}\\     
    \bottomrule
    \end{tabular}
\caption{Experimental results over the MeasEval Benchmark. Comparing MeasHalu with competition leaders and rule/sentence-based LLM baselines. Top ranks are shaded orange (1st), yellow (2nd), and teal (3rd).}
\label{tab:relation-mitigation}
\end{table*}

\begin{table*}[!htp]
\centering
\setlength{\tabcolsep}{1pt} 
\scriptsize
\resizebox{\linewidth}{!}{
    \begin{tabular}{l c c ccc c cc c cc c cc} 
    \toprule
    \multirow{3}{*}{\textbf{Model}} & \multirow{3}{*}{\textbf{Overall}} & & \multicolumn{3}{c}{\textbf{Quantities}} & & \multicolumn{2}{c}{\textbf{Entities}} & & \multicolumn{2}{c}{\textbf{Properties}} & & \multicolumn{2}{c}{\textbf{Qualifiers}} \\
    \cmidrule{4-6}\cmidrule{8-9}\cmidrule{11-12}\cmidrule{14-15}
     & & & \textbf{Quantity} & \textbf{Unit} & \textbf{Modifier} & & \textbf{ME} & \textbf{HasQuantity} & & \textbf{MP} & \textbf{HasProperty} & & \textbf{Qualifier} & \textbf{Qualifies} \\
    \midrule

    \multicolumn{15}{c}{\textit{Rule-based Prompting}} \\ 
    \midrule
    
    GPT-5 & \valueWithStd{0.383}{0.004} && \cellcolor{lightyellow}{\valueWithStd{0.833}{0.021}} & \valueWithStd{0.706}{0.017} & \valueWithStd{0.357}{0.031} && \valueWithStd{0.400}{0.004} & \valueWithStd{0.282}{0.003} && \cellcolor{lightteal}{\valueWithStd{0.310}{0.020}} & \valueWithStd{0.046}{0.011} && \cellcolor{lightteal}{\valueWithStd{0.135}{0.017}} & \valueWithStd{0.019}{0.012} \\
    DeepSeek-R1 & \valueWithStd{0.252}{0.006} && \valueWithStd{0.553}{0.016} & \valueWithStd{0.514}{0.046} & \valueWithStd{0.217}{0.018} && \valueWithStd{0.213}{0.005} & \valueWithStd{0.169}{0.009} && \valueWithStd{0.141}{0.015} & \valueWithStd{0.045}{0.006} && \valueWithStd{0.099}{0.009} & \valueWithStd{0.052}{0.018} \\
    DeepSeek-V3 & \valueWithStd{0.312}{0.003} && \valueWithStd{0.724}{0.014} & \valueWithStd{0.726}{0.011} & \valueWithStd{0.234}{0.009} && \valueWithStd{0.270}{0.014} & \valueWithStd{0.189}{0.003} && \valueWithStd{0.121}{0.005} & \valueWithStd{0.012}{0.003} && \valueWithStd{0.113}{0.018} & \valueWithStd{0.037}{0.006} \\
    Gemini-2.5-Pro & \valueWithStd{0.386}{0.009} && \valueWithStd{0.766}{0.005} & \valueWithStd{0.707}{0.012} & \valueWithStd{0.444}{0.026} && \valueWithStd{0.351}{0.014} & \cellcolor{lightteal}{\valueWithStd{0.312}{0.014}} && \valueWithStd{0.291}{0.028} & \valueWithStd{0.151}{0.017} && \cellcolor{lightyellow}{\valueWithStd{0.140}{0.017}} & \valueWithStd{0.055}{0.006} \\
    Qwen2.5-72b & \valueWithStd{0.296}{0.007} && \valueWithStd{0.675}{0.012} & \valueWithStd{0.747}{0.010} & \valueWithStd{0.203}{0.025} && \valueWithStd{0.246}{0.015} & \valueWithStd{0.187}{0.015} && \valueWithStd{0.142}{0.004} & \valueWithStd{0.000}{0.000} && \valueWithStd{0.091}{0.014} & \valueWithStd{0.066}{0.009} \\
    Qwen2.5-7b & \valueWithStd{0.181}{0.009} && \valueWithStd{0.501}{0.016} & \valueWithStd{0.353}{0.031} & \valueWithStd{0.148}{0.037} && \valueWithStd{0.108}{0.015} & \valueWithStd{0.078}{0.010} && \valueWithStd{0.069}{0.012} & \valueWithStd{0.001}{0.001} && \valueWithStd{0.038}{0.025} & \valueWithStd{0.006}{0.004} \\

    \midrule
    \multicolumn{15}{c}{\textit{Sentence-based Prompting}} \\ 
    \midrule
    
    GPT-5 & \cellcolor{lightteal}{\valueWithStd{0.402}{0.007}} && \valueWithStd{0.750}{0.002} & \cellcolor{lightteal}{\valueWithStd{0.759}{0.005}} & \cellcolor{lightyellow}{\valueWithStd{0.458}{0.007}} && \cellcolor{lightteal}{\valueWithStd{0.435}{0.013}} & \valueWithStd{0.303}{0.007} && \valueWithStd{0.303}{0.010} & \cellcolor{lightteal}{\valueWithStd{0.239}{0.018}} && \valueWithStd{0.100}{0.016} & \cellcolor{lightteal}{\valueWithStd{0.070}{0.011}} \\
    DeepSeek-R1 & \valueWithStd{0.324}{0.013} && \valueWithStd{0.622}{0.021} & \valueWithStd{0.648}{0.018} & \valueWithStd{0.255}{0.012} && \valueWithStd{0.299}{0.014} & \valueWithStd{0.235}{0.019} && \valueWithStd{0.215}{0.008} & \valueWithStd{0.146}{0.013} && \valueWithStd{0.075}{0.015} & \valueWithStd{0.069}{0.014} \\
    DeepSeek-V3 & \valueWithStd{0.299}{0.012} && \valueWithStd{0.521}{0.023} & \valueWithStd{0.565}{0.020} & \valueWithStd{0.229}{0.007} && \valueWithStd{0.296}{0.017} & \valueWithStd{0.229}{0.012} && \valueWithStd{0.209}{0.016} & \valueWithStd{0.155}{0.008} && \valueWithStd{0.053}{0.004} & \valueWithStd{0.048}{0.004} \\
    Gemini-2.5-Pro & \cellcolor{lightyellow}{\valueWithStd{0.462}{0.007}} && \cellcolor{lightteal}{\valueWithStd{0.827}{0.009}} & \cellcolor{lightred}{\valueWithStd{0.832}{0.007}} & \cellcolor{lightteal}{\valueWithStd{0.444}{0.010}} && \cellcolor{lightyellow}{\valueWithStd{0.472}{0.014}} & \cellcolor{lightyellow}{\valueWithStd{0.399}{0.012}} && \cellcolor{lightyellow}{\valueWithStd{0.402}{0.006}} & \cellcolor{lightyellow}{\valueWithStd{0.332}{0.015}} && \valueWithStd{0.100}{0.002} & \cellcolor{lightyellow}{\valueWithStd{0.072}{0.003}} \\
    Qwen2.5-72b & \valueWithStd{0.202}{0.006} && \valueWithStd{0.343}{0.008} & \valueWithStd{0.383}{0.006} & \valueWithStd{0.183}{0.007} && \valueWithStd{0.207}{0.012} & \valueWithStd{0.145}{0.012} && \valueWithStd{0.127}{0.007} & \valueWithStd{0.106}{0.010} && \valueWithStd{0.048}{0.001} & \valueWithStd{0.050}{0.007} \\
    Qwen2.5-7b & \valueWithStd{0.033}{0.003} && \valueWithStd{0.065}{0.002} & \valueWithStd{0.056}{0.005} & \valueWithStd{0.022}{0.005} && \valueWithStd{0.030}{0.001} & \valueWithStd{0.028}{0.006} && \valueWithStd{0.017}{0.005} & \valueWithStd{0.008}{0.003} && \valueWithStd{0.016}{0.007} & \valueWithStd{0.020}{0.007} \\
        
    \midrule
    MeasHalu-7B & \cellcolor{lightred}{\valueWithStd{0.578}{0.002}} && \cellcolor{lightred}{\valueWithStd{0.861}{0.003}} & \cellcolor{lightyellow}{\valueWithStd{0.832}{0.012}} & \cellcolor{lightred}{\valueWithStd{0.539}{0.005}} && \cellcolor{lightred}{\valueWithStd{0.555}{0.006}} & \cellcolor{lightred}{\valueWithStd{0.551}{0.008}} && \cellcolor{lightred}{\valueWithStd{0.522}{0.006}} & \cellcolor{lightred}{\valueWithStd{0.459}{0.007}} && \cellcolor{lightred}{\valueWithStd{0.159}{0.021}} & \cellcolor{lightred}{\valueWithStd{0.097}{0.004}} \\

    \bottomrule
    \end{tabular}
}
\caption{Experimental results over the MeasEval-Ext.}
\label{tab:results_self_dataset_with_std}
\end{table*}

\begin{table*}[t]
\centering
\setlength{\tabcolsep}{3pt}
\footnotesize
    \begin{tabular}{l l c c c c c c c} 
    \toprule
    \multirow{2.5}{*}{\textbf{Inference Model}} & \multirow{2.5}{*}{\textbf{Context Source}} & \multirow{2.5}{*}{\textbf{Validity}} & \multicolumn{2}{c}{\textbf{BLEU}} & \textbf{LEV} & \multicolumn{3}{c}{\textbf{ROUGE}} \\
    \cmidrule{4-5} \cmidrule{7-9}
     & & & \textbf{B-2} & \textbf{B-4} & \textbf{50\%} & \textbf{R-1} & \textbf{R-2} & \textbf{R-L} \\
    \midrule

    \multirow{3}{*}{Gemini-2.5-Pro} 
      & MeasHalu & \textbf{\valueWithStd{14.67}{2.31}} & \textbf{\valueWithStd{58.72}{0.23}} & \textbf{\valueWithStd{44.23}{0.29}} & \textbf{\valueWithStd{60.00}{2.00}} & \textbf{\valueWithStd{71.71}{0.18}} & \textbf{\valueWithStd{54.24}{0.32}} & \textbf{\valueWithStd{66.51}{0.26}} \\
      & Gemini   & \valueWithStd{12.67}{1.53} & \valueWithStd{58.26}{0.60} & \valueWithStd{43.99}{0.67} & \valueWithStd{59.67}{3.06} & \valueWithStd{71.28}{0.39} & \valueWithStd{54.21}{0.55} & \valueWithStd{66.49}{0.44} \\
      & None     & \valueWithStd{0.33}{0.58}  & \valueWithStd{51.78}{0.32} & \valueWithStd{37.23}{0.36} & \valueWithStd{37.33}{6.66} & \valueWithStd{62.92}{0.39} & \valueWithStd{46.25}{0.27} & \valueWithStd{58.90}{0.26} \\
    \midrule
    
    \multirow{3}{*}{DeepSeek-R1} 
      & MeasHalu & \textbf{\valueWithStd{10.67}{1.53}} & \textbf{\valueWithStd{58.99}{0.37}} & \textbf{\valueWithStd{43.01}{0.56}} & \textbf{\valueWithStd{61.33}{2.89}} & \textbf{\valueWithStd{71.62}{0.19}} & \textbf{\valueWithStd{52.79}{0.12}} & \textbf{\valueWithStd{65.85}{0.40}} \\
      & Gemini   & \valueWithStd{10.33}{0.58} & \valueWithStd{58.28}{0.73} & \valueWithStd{42.37}{0.61} & \valueWithStd{56.67}{2.89} & \valueWithStd{71.13}{0.38} & \valueWithStd{52.31}{0.36} & \valueWithStd{65.65}{0.31} \\
      & None     & \valueWithStd{8.67}{1.15}  & \valueWithStd{38.32}{1.02} & \valueWithStd{26.21}{0.99} & \valueWithStd{22.00}{5.29} & \valueWithStd{59.12}{0.31} & \valueWithStd{42.32}{0.21} & \valueWithStd{53.56}{0.41} \\
        \midrule

    \multirow{3}{*}{GPT-5} 
      & MeasHalu & \textbf{\valueWithStd{16.33}{2.52}} & \textbf{\valueWithStd{51.55}{0.29}} & \valueWithStd{37.43}{0.52} & \textbf{\valueWithStd{52.33}{3.51}} & \textbf{\valueWithStd{71.00}{0.39}} & \valueWithStd{51.62}{0.27} & \textbf{\valueWithStd{65.47}{0.32}} \\
      & Gemini   & \valueWithStd{13.33}{2.08} & \valueWithStd{50.61}{0.72} & \valueWithStd{36.56}{0.81} & \valueWithStd{48.33}{4.04} & \valueWithStd{70.73}{0.80} & \valueWithStd{51.31}{0.64} & \valueWithStd{65.06}{0.66} \\
      & None     & \valueWithStd{1.35}{1.54}  & \valueWithStd{50.39}{0.58} & \textbf{\valueWithStd{39.39}{0.78}} & \valueWithStd{50.17}{2.33} & \valueWithStd{66.31}{0.35} & \textbf{\valueWithStd{52.53}{0.40}} & \valueWithStd{62.76}{0.44} \\
    
    \bottomrule
    \end{tabular}
\caption{Performance comparison on the OpenExp-Action-100 dataset. Models are provided with \textbf{MeasEval-formatted quantities and relations} generated by different sources (MeasHalu vs. Gemini) as context. The best scores \textbf{for each model} are highlighted in \textbf{bold}.}
\label{tab:reactxt_results}
\end{table*}

\begin{figure*}[ht]
\includegraphics[width=\textwidth]{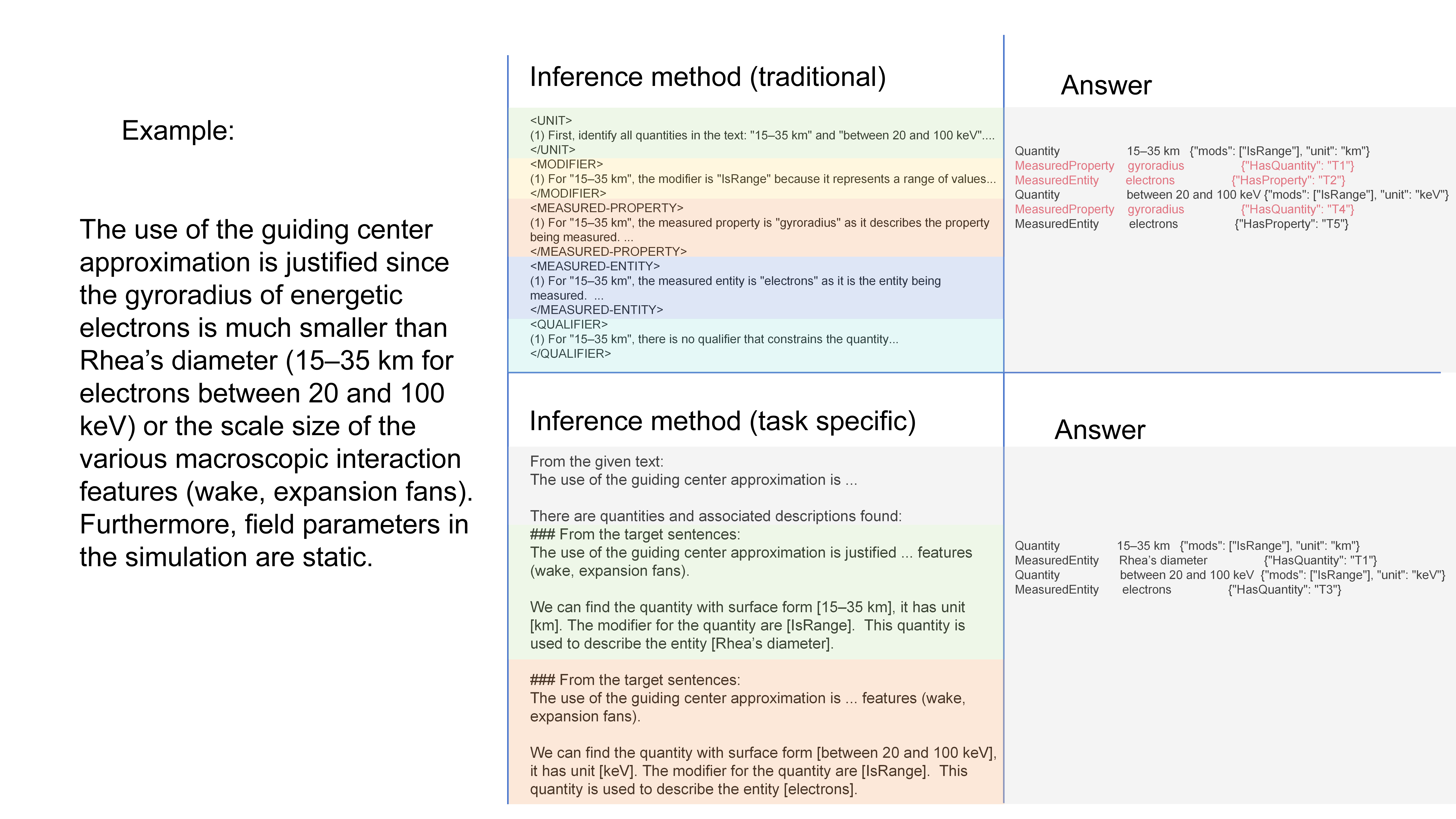}
 \caption{Comparison of sentence-based and rule-based reasoning approaches}
\end{figure*}

\section{Implementation Details}

\subsection{Supervised Fine-Tuning (SFT)}

For the Supervised Fine-Tuning (SFT) stage in MeasHalu, we adopt the LlamaFactory~\cite{zheng2024llamafactory} framework for model training. When applying parameter-efficient fine-tuning with Low-Rank Adaptation (LoRA), the training is conducted on a single NVIDIA A800 GPU. In contrast, full-parameter fine-tuning requires increased computational resources and is therefore performed using two NVIDIA A800 GPUs.

\subsection{GRPO Training}

For the GRPO stage, we utilize the Unsloth framework~\cite{unsloth} to improve training efficiency. The entire GRPO training process is carried out on a single NVIDIA A800 GPU.

\end{CJK*}
\end{document}